\documentclass[11pt]{article}

\usepackage[final]{acl}

\usepackage{times}
\usepackage{latexsym}

\usepackage[T1]{fontenc}

\usepackage[utf8]{inputenc}

\usepackage{microtype}

\usepackage{inconsolata}

\usepackage{graphicx}

\usepackage[most]{tcolorbox}
\usepackage{booktabs}
\usepackage{amsmath}
\usepackage{amsfonts}
\usepackage{tabularx}
\usepackage{multicol}
\usepackage{multirow}
\usepackage{makecell}
\usepackage{hyperref}
\usepackage{xurl}
\usepackage[ruled, linesnumbered]{algorithm2e}
\usepackage{enumitem}
\usepackage{soul}
\usepackage[table]{xcolor}
\definecolor{hallEntity}{HTML}{C0392B}        
\definecolor{hallNumber}{HTML}{E67E22}        
\definecolor{hallFalseConcat}{HTML}{8E44AD}   
\definecolor{hallContext}{HTML}{2874A6}       
\definecolor{hallAttribution}{HTML}{16A085}   
\definecolor{hallReasoning}{HTML}{B7950B}     
\definecolor{hallOvergen}{HTML}{1E8449}       
\definecolor{hallHyperbole}{HTML}{C2185B}     
\definecolor{hallTemporal}{HTML}{283593}      
\definecolor{hallNA}{HTML}{95A5A6}            

\definecolor{paperbar}{HTML}{ECEFF1}          

\newcommand{\halltag}[2]{%
  {\setlength{\fboxsep}{2.5pt}%
   \colorbox{#1}{\color{white}\scriptsize\sffamily\bfseries\,#2\,}}%
}
\newcommand{\hentity}{\halltag{hallEntity}{Entity}}
\newcommand{\hnumber}{\halltag{hallNumber}{Number}}
\newcommand{\hfalseconcat}{\halltag{hallFalseConcat}{False Concatenation}}
\newcommand{\hcontext}{\halltag{hallContext}{Context-based Meaning}}
\newcommand{\hattribution}{\halltag{hallAttribution}{Attribution Failure}}
\newcommand{\hreasoning}{\halltag{hallReasoning}{Reasoning Error}}
\newcommand{\hovergen}{\halltag{hallOvergen}{Overgeneralization}}
\newcommand{\hhyperbole}{\halltag{hallHyperbole}{Hyperbole}}
\newcommand{\htemporal}{\halltag{hallTemporal}{Temporal}}

\newcommand{\paperrow}[1]{%
  \rowcolor{paperbar}\multicolumn{3}{@{}l@{}}{%
    \footnotesize\sffamily\textbf{Paper:}~\textit{#1}%
  }\\[1pt]%
}

\newcommand{\keyquote}[1]{%
  \par\smallskip\noindent
  {\scriptsize\color{gray!30!black}\sffamily\textbf{Key quote:}~%
   \normalfont\itshape``#1''}%
}

\definecolor{changes}{RGB}{200, 255, 220} 
\definecolor{todo}{RGB}{255, 200, 220} 
\definecolor{prev}{RGB}{241, 225, 255} 

\soulregister{\cite}{7}
\soulregister{\citep}{7}
\soulregister{\noindent}{7}
\soulregister{\textbf}{7}
\soulregister{\citet}{7}
\soulregister{\ref}{7}
\soulregister{\pageref}{7}
\usepackage[htt]{hyphenat}

\usepackage{dirtree}
\definecolor{depthZero}{HTML}{9C27B0}  
\definecolor{depthOne}{HTML}{2196F3}   
\definecolor{depthTwo}{HTML}{00BCD4} 
\definecolor{depthThree}{HTML}{4CAF50} 
\definecolor{descColor}{HTML}{757575}  

\SetKwInOut{Parameter}{parameter}

\title{HalluPeer: A Taxonomy-driven Benchmark for Detecting Hallucinations in Scientific Peer Reviews}

\author{
Tzu-Ling Lin, 
Dong-Ting Yao, 
Teng-Fang Hsiao, 
Wei-Chih Chen,
Hong-Han Shuai\thanks{Corresponding author.} \\
National Yang Ming Chiao Tung University \\
tzulinglin.11@nycu.edu.tw}

\begin{document}
\maketitle
\begin{abstract}
The growing scale of academic peer review has motivated the use of Large Language Models (LLMs) as review assistants, yet LLMs can generate fluent but unsupported claims that undermine review reliability. Existing hallucination benchmarks are not designed for peer review, where verification requires grounding claims in long, technical papers. We introduce \textbf{HalluPeer}, a benchmark for detecting hallucinations in scientific peer reviews, providing aligned triples of paper content, human-written reviews, and hallucination-injected reviews, annotated for detection, classification, and localization. Our pipeline induces a peer-review-specific hallucination taxonomy, identifies review contexts, and injects hallucinations with automated filtering. Experiments on 12K papers and 38K reviews show that existing detectors struggle to separate hallucinations from legitimate critique, while evaluation on authentic reviews demonstrates that HalluPeer-defined hallucination patterns occur in real peer reviews, highlighting the critical need for source-aware verification. Our project page can be found in \url{https://github.com/Lin-TzuLing/HalluPeer.git}
\end{abstract}

\section{Introduction}

\begin{figure*}
    \centering
    \includegraphics[width=\linewidth]{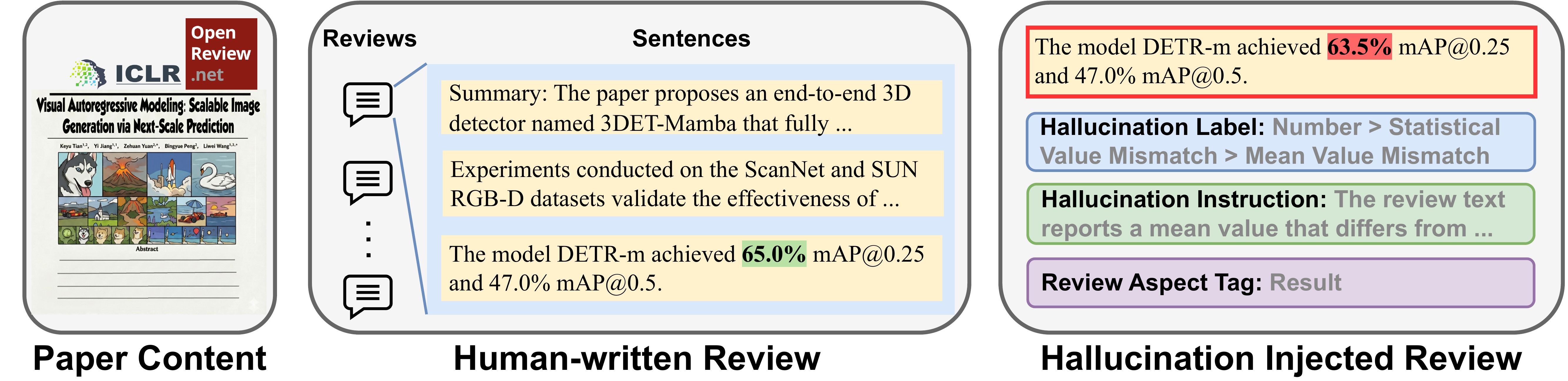}
    \caption{\textbf{HalluPeer benchmark Overview.} The benchmark comprises triples of paper content, original human-written reviews, and hallucinated reviews. Each triple is enriched with hallucination types, injection instructions, and aspect tags to facilitate systematic benchmarking of hallucination detection in peer reviews.}
    \label{fig:intro}
\end{figure*}

The rapid growth of AI research increasingly burdens the peer-review system: submissions to major NLP conferences like ACL and EMNLP have surged from approximately 3.4K in 2020 to over 8K in 2025, making review quality, consistency, and timeliness difficult to maintain. To address this pressure, AI-assisted peer review has emerged as a promising solution. LLMs are increasingly used to draft comments, structure critiques, and streamline meta-reviews~\citep{LLMReview_ou2025claimcheck, yu2024reviewedbyllm}; moving beyond prototypes, conferences like AAAI 2026 have piloted LLM integration for initial reviews and committee summarizations,\footnote{\url{https://aaai.org/aaai-launches-ai-powered-peer-review-assessment-system/}} aiming to boost efficiency while retaining human oversight.

However, AI-assisted reviewing introduces a critical reliability problem: \emph{hallucination in paper reviews}. Despite careful prompting and system safeguards, LLMs can generate fluent but unsupported claims, \textit{e.g.}, falsely asserting a missing baseline, misreporting results, or fabricating assumptions. These factual errors transcend stylistic flaws; they mislead meta-reviewers and unfairly impact editorial decisions. Detecting them is therefore essential for trustworthy AI-assisted review workflows.

However, detecting these hallucinations is harder than in settings where the problem has been widely studied, such as QA, summarization, and RAG~\citep{HalluDetect_Hades_liu2022,HalluDetect_SelfCheckGPT_manakul2023,HalluDetect_absSum_maynez2020,HalluDetect_HaluEval_li2023,HalluDetect_RAG_niu2024ragtruth}. Verifying a review requires synthesizing evidence across long, technical sections (e.g., methods, tables, appendices), and because reviews intertwine facts with subjective critiques, a detector must verify paper-grounded claims while setting aside evaluative opinions. These demands also explain why the problem remains under-studied: existing hallucination datasets rarely provide the annotation structure this domain needs—paper-grounded evidence for checking review claims, and fine-grained hallucination types reflecting scientific critique. Without these dimensions, evaluation cannot reveal whether a model fails because it cannot retrieve the right evidence, reason across sections, or recognize a specific error type such as a wrong number, fabricated comparison, or unsupported attribution.

Thus, we introduce \textbf{HalluPeer}, a taxonomy-driven benchmark for detecting hallucinations in scientific peer reviews. We formulate review hallucination detection as a paper-grounded verification problem: \textbf{a review claim is hallucinated if it is unsupported by, or incorrect with respect to, the submitted paper}. This strictly isolates factual grounding from tone and subjective judgments.

As shown in Fig.~\ref{fig:intro}, HalluPeer contains aligned triples of \emph{paper content}, \emph{human-written reviews}, and \emph{hallucination-injected reviews}. Construction is guided by two principles: \textbf{coverage} of scientific failure modes, and \textbf{control} over error types and contexts. To achieve this, we induce a peer-review-specific hierarchical hallucination taxonomy, segment human-written reviews into sentences, assign review-aspect tags, and construct hallucination templates by pairing taxonomy instructions with aspect-compatible contexts. A constrained LLM editor then introduces hallucinations while preserving style, followed by automated filtering. 

Based on HalluPeer, we evaluate existing verifiers across three tasks: \textbf{detection, classification, and localization}. Results indicate that current methods struggle with scientific reviews. General verifiers often confuse unsupported claims with legitimate critique, while LLM judges fail on errors requiring technical grounding. An analysis on authentic reviews shows that HalluPeer-defined hallucination patterns occur in real peer reviews, and a detector trained solely on HalluPeer recovers all expert-annotated hallucinations in authentic reviews, while source attribution remains a challenge. Overall, these results suggest that auditing scientific reviews requires specialized, source-aware verification rather than off-the-shelf factuality models.

Our contributions are summarized as follows:
\begin{itemize}[leftmargin=*, noitemsep, topsep=0pt, parsep=0pt]
    \item We introduce \textbf{HalluPeer}, a benchmark consisting of paper, human-review, and hallucination-injected-review, with annotations for detection, type classification, and localization.
    \item We propose a taxonomy-driven and aspect-conditioned construction pipeline that induces peer-review-specific hallucination types, identifies compatible review contexts, and injects hallucinations with semantic verification.
    \item We systematically evaluate hallucination detection, classification, and localization, showing that general-purpose verifiers struggle with scientific reviews while domain-specific fine-tuning substantially improves performance. Additional cross-generator, cross-venue, and authentic-review evaluations demonstrate robustness beyond the original synthetic setting and transfer to naturally occurring reviewer errors.
\end{itemize}


\section{Related Work}
\subsection{LLM Hallucination Detection}
Hallucination detection identifies generated statements unsupported by evidence, and has been studied in question answering, summarization, and retrieval-augmented generation. Some methods operate without gold references, using model output or consistency across sampled responses~\citep{HalluDetect_Hades_liu2022,HalluDetect_SelfCheckGPT_manakul2023}, while others define hallucination relative to explicit evidence such as source documents~\citep{HalluDetect_absSum_maynez2020,HalluDetect_absSum_cao2022,HalluDetect_absSum_bao2025faithbench} or retrieved passages~\citep{HalluDetect_RAG_sriramanan2024,HalluDetect_RAG_niu2024ragtruth}. Recent benchmarks add labels, span attribution, and type-level diagnosis~\citep{HalluDetect_FAVAbench_mishra2024,HalluDetect_HalluMeasure_akbar2024,HalluDetect_HalluLens_bang2025}. However, these resources are built on general-domain text or retrieved snippets, and do not capture the verification structure of scientific peer review, where evidence is a long technical paper and a review claim may require reasoning across various sections. HalluPeer addresses this gap with paper-aligned evidence and review-specific hallucination.
\begin{figure*}[h]
    \centering
    \includegraphics[width=\linewidth]{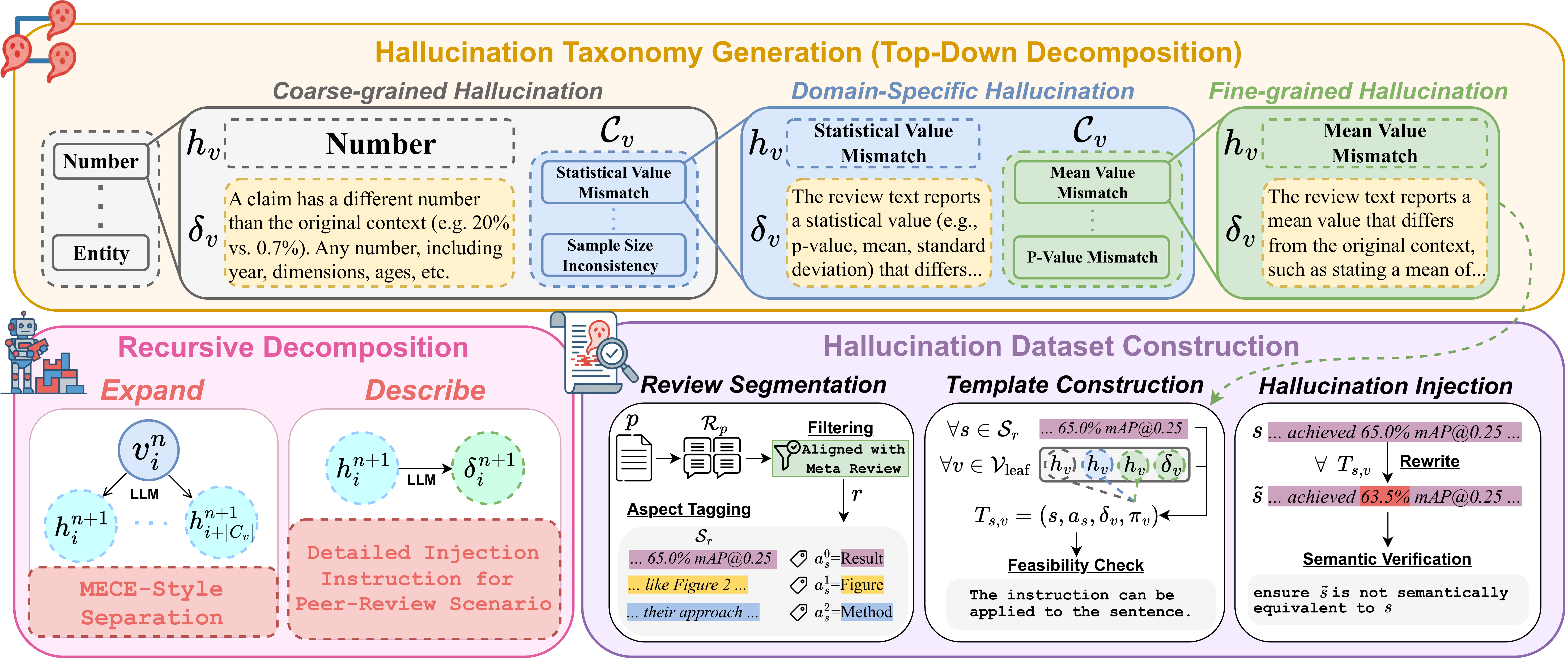}
    \caption{\textbf{Overview of HalluPeer construction framework}. We build a top-down hallucination taxonomy for peer-review scenario and derive fine-grained instructions. These instructions are used to construct sentence-level injection templates and generate hallucinated review sentences via an automated injection and verification pipeline.}
    \label{fig:method}
\end{figure*}
\subsection{LLMs in Automated Peer Review}
The growing reviewing burden has motivated the use of LLMs for review drafting, structured critique, meta-review assistance, and review-quality assessment~\citep{Survey_AIReview_russo2025,Survey_AIReview_thakkar2025,Survey_AIReview_zhuang2025,LLMReview_du2024}. However, factual grounding remains a challenge: LLM reviewers may produce detailed feedback for incomplete manuscripts~\citep{LLMReview_ye2024} and generate inconsistent or paper-unsupported claims~\citep{LLMReview_du2024,LLMReview_ou2025claimcheck}.
These findings establish review hallucination as a realistic risk in scientific peer review. Nevertheless, prior work lacks a dedicated benchmark for systematically detecting, categorizing, and localizing hallucinated review claims grounded in the submitted paper. HalluPeer is designed to address this gap.

\section{Taxonomy of Review Hallucinations}
\label{sec:taxonomy_of_peer_review_hallucinations}
Existing hallucination taxonomies are derived from general-domain text and rely on high-level distinctions, \textit{e.g.}, factual vs. faithfulness or intrinsic vs. extrinsic hallucinations~\cite{CoarseTaxonomy_ji2023survey, CoarseTaxonomy_li2024dawn}. Such categories are insufficient for peer reviews, where unsupported claims often involve paper-specific numbers or attribution errors. As shown in Fig.~\ref{fig:method}, we refine these coarse categories into a hierarchical taxonomy of fine-grained review hallucination types, along with operational instructions for controlled injection and evaluation. Sec.~\ref{subsec:top_down_taxonomy_construction} and~\ref{subsec:LLM_guided_recursive_decomposition} describe the taxonomy structure and decomposition procedure.



\subsection{Top-Down Taxonomy Construction}
\label{subsec:top_down_taxonomy_construction}

Our benchmark requires hallucination labels that are both \textit{controllable} for data construction and \textit{diagnostic} for evaluation. To this end, we construct a hierarchical, tree-structured taxonomy in a top-down manner, where coarse categories are recursively refined into fine-grained, review-specific error types.

\noindent\textbf{Taxonomy structure.} We represent the taxonomy as a rooted tree over a set of nodes $\mathcal{V}$. Each node $v \in \mathcal{V}$ corresponds to a hallucination concept and is defined as $v=\langle h_v, \delta_v, p_v, \mathcal{C}_v \rangle$, where $h_v$ is the name of the hallucination concept, $\delta_v$ is an operational description, $p_v$ is the parent node, and $\mathcal{C}_v = \{ u \in \mathcal{V} \mid p_u = v \}$ is the set of child nodes. Aggregating along root-to-leaf paths supports analysis at different granularity levels. For each leaf node $v \in \mathcal{V}_{\text{leaf}}$ (where $\mathcal{C}_v=\emptyset$), the description $\delta_v$ gives an explicit specification of how the hallucination should appear in review text (\textit{e.g.}, what to alter, fabricate, or violate); these leaf specifications serve both as controllable injection instructions and as fine-grained evaluation labels. 

\noindent\textbf{Initialization with coarse anchors.} We initialize the first level with a set of coarse-grained semantic anchors adapted from prior hallucination categorization work \citep{HalluDetect_HalluMeasure_akbar2024}: nine categories capturing common forms of unsupported generation, \textit{i.e.}, \textit{Number}, \textit{Entity}, \textit{False Concatenation}, \textit{Attribution Failure}, \textit{Overgeneralization}, \textit{Reasoning Error}, \textit{Hyperbole}, \textit{Temporal}, and \textit{Context-based Meaning Error}.\footnote{Definitions are provided in the Appendix. We omit the \textit{Other} category due to its incompatibility for injection.} These anchors allow downstream refinement into review-specific manifestations.

\subsection{LLM-Guided Recursive Decomposition}
\label{subsec:LLM_guided_recursive_decomposition}

We expand each coarse anchor into review-specific subtypes via a depth-first generation procedure (Algorithm~\ref{alg:taxonomy_decomposition}). At each recursive step, an LLM proposer ($\mathcal{M}$) is queried to generate candidate child nodes. The recursion designates a node as a terminal leaf under two conditions: reaching a predefined maximum depth, or yielding no further valid subdivisions from the LLM (empty expansion).

\noindent\textbf{Two proposer functions.} The proposer $\mathcal{M}$ is used through two functions. $\textsc{Expand}_{\mathcal{M}}$ takes a parent concept and its description and returns a list of child concepts; $\textsc{Describe}_{\mathcal{M}}$ takes a child concept and produces an operational definition of how that hallucination manifests in review text. Each child is then attached to its parent, added to the global node set, assigned its description, and recursively decomposed at the next depth.

\noindent\textbf{Constraints for usable subcategories.}
To keep the taxonomy usable for fine-grained labeling and controllable injection, we enforce several constraints through prompting (templates in Appendix~\ref{app: prompt_hallucination_taxonomy_generation}). First, generated children should follow a Mutually Exclusive and Collectively Exhaustive (MECE)-style separation: sibling categories should be minimally overlapping while collectively covering the major manifestations of the parent concept. Second, children should be grounded in concrete peer-review scenarios (\textit{e.g.}, claims about baselines or experiments), avoiding abstract distinctions. Third, the proposer keeps sibling categories at comparable granularity and returns an empty list when the parent is already atomic at the current depth.

\begin{figure}[t]
    \centering
    \includegraphics[width=\linewidth]{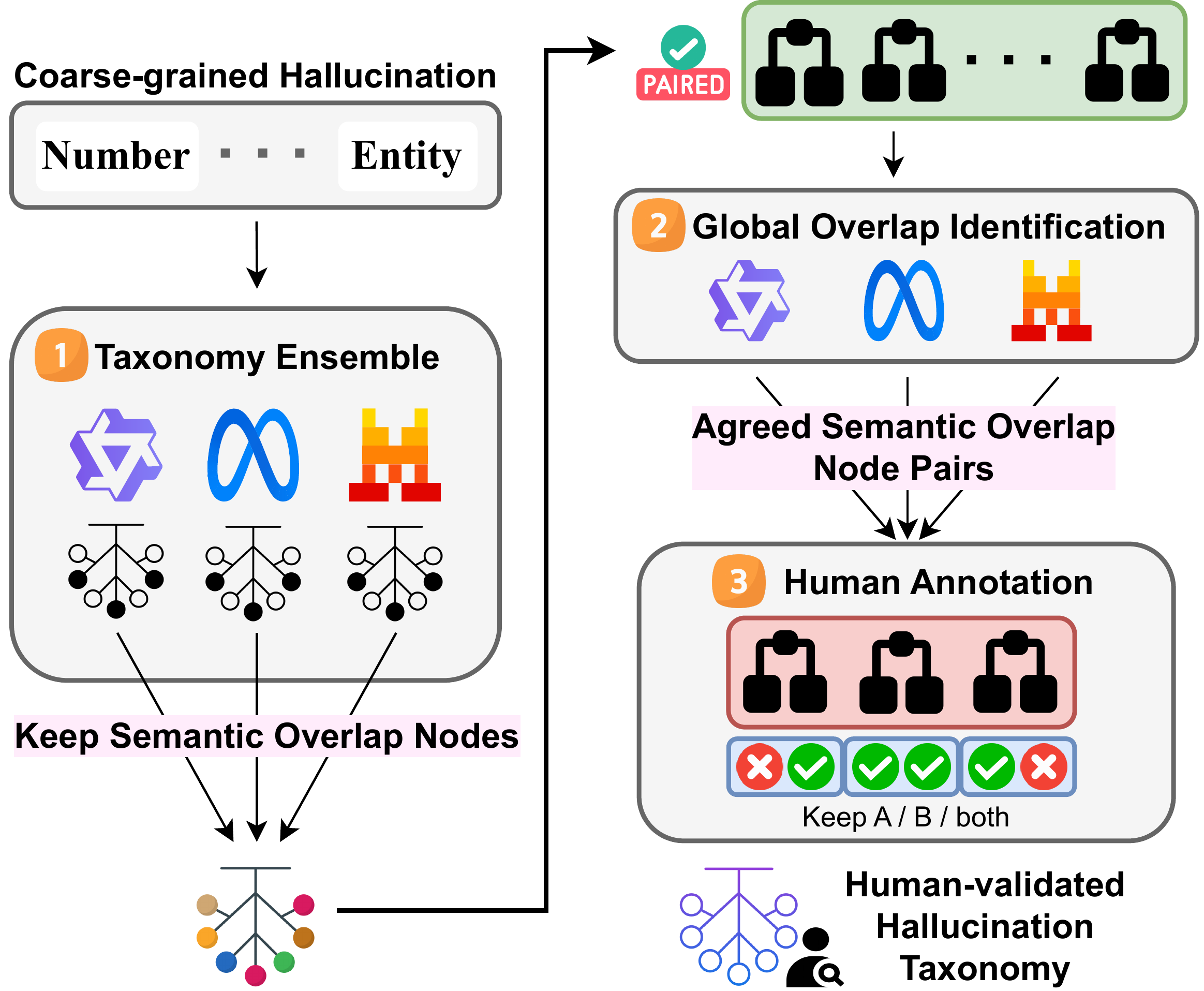}
    \caption{\textbf{Overview of the multi-model ensemble and expert refinement pipeline.} The workflow consists of (1) generating taxonomy trees using multiple LLM proposers, (2) identifying globally overlapping concepts through multi-LLM consensus filtering, and (3) performing expert validation to finalize the taxonomy.}
    \label{fig:ensemble_pipeline}
    \vspace{-3mm}
\end{figure}

\subsection{Ensemble and Expert Refinement}
\label{subsec:ensemble_and_expert_refinement}
Relying on a single LLM proposer raises two risks: generative priors may overproduce idiosyncratic concepts, and independently decomposed branches may introduce global semantic redundancy across the hierarchy. 
We address both via a three-stage workflow (Fig.~\ref{fig:ensemble_pipeline}): (1) multi-model ensemble with cross-tree agreement filtering, (2) global overlap identification, and (3) human expert validation. Stage 1 first retains taxonomy nodes with cross-model consensus, while stage 2 performs pairwise overlap identification among the retained nodes. During this stage, a multi-LLM consensus process reduces 86,800 candidate node pairs to 434 high-overlap pairs. At stage 3, human experts then inspect to decide whether concepts should be merged or kept distinct, ensuring that the taxonomy remains conceptually coherent and distinguishable. The final taxonomy contains 265 nodes after refinement. Full details are provided in Appendix~\ref{app:ensemble_details}.

\section{HalluPeer Construction Pipeline}
\subsection{Data Collection}
\label{subsec:data_collection}
We source ICLR (2019–2024) and NeurIPS (2021–2024) records from OpenReview, as their large-scale submissions and reviews are crucial for paper-grounded hallucination construction and validation (see Appendix~\ref{app:data_stats} for statistics).

Following prior work on review aspect analysis~\citep{ReviewAspectTag_lu2025identifying}, we tag each review sentence with an aspect label (e.g., \emph{Novelty}, \emph{Evaluation}, \emph{Clarity}) and use it as an injection constraint: the injected hallucination type must remain compatible with the sentence's role. For instance, number-related perturbations are preferentially injected into \emph{Evaluation} sentences and avoided in \emph{Clarity} sentences, where such edits would seem unnatural.

Although peer reviews are human-written, they may still contain unsupported statements, which would make them unreliable sources for hallucination injection. While meta-reviews are not factual ground truth, they provide a useful proxy for review points that were salient to the final assessment. We therefore use an LLM filter $\textsc{Filter}_{\mathcal{M}}$ to compare each review with the corresponding meta-review and assign an alignment judgment with confidence. We use this alignment as a conservative selection heuristic: only high-confidence, meta-review-aligned reviews are retained for hallucination injection, reducing the potential risk of selecting unreliable base reviews.

\subsection{Injection Template Construction}
\label{subsec:hallucination_template_construction}
We construct hallucination injection templates by pairing review sentences with fine-grained hallucination concepts from the taxonomy. Each review $r \in \mathcal{R}_p$ is segmented into sentences $\mathcal{S}_r = \{s_1, \ldots, s_{|\mathcal{S}_r|}\}$, where each $s \in \mathcal{S}_r$ is annotated with an aspect label $a_s$. (Sec.~\ref{subsec:data_collection}). On the taxonomy side, each leaf node $v \in \mathcal{V}_{\text{leaf}}$ corresponds to a fine-grained concept specified by an injection instruction $\delta_v$ and a label $\pi_v$ given by its root-to-$v$ path. Pairing a sentence $s$ with a leaf node $v$ then yields a hallucination injection template
\[
T_{s,v} = (s, a_s, \delta_v, \pi_v).
\]

Naively instantiating every $(s, v)$ pair is intractable, so we narrow the search space by first applying a coarse-grained screening: for each sentence we test which first-level concepts are even applicable (e.g., \emph{Number} requires explicit numeric values), and build fine-grained templates only from the leaf nodes descending from the compatible anchors. Finally, an LLM-based compatibility check $\textsc{Check}_{\mathcal{M}}$ filters out templates whose hallucination type cannot be naturally applied to the target sentence and aspect, yielding the feasible set $\mathcal{T}^{\text{feasible}}$. The screening prompt is in Appendix~\ref{app: prefilter}, and the pseudocode is Appendix~\ref{app:algorithm}.

\subsection{Automated Injection Pipeline}
Given a set of feasible templates, we generate hallucinated counterparts of human-written sentences via an automated pipeline. For each $T_{s,v} \in \mathcal{T}^{\text{feasible}}$, we prompt an LLM injector $\textsc{Inject}_{\mathcal{M}}$ with the original sentence $s$ and the taxonomy instruction $\delta_v$, producing a hallucinated sentence $\tilde{s}=\textsc{Inject}_{\mathcal{M}}(s,\delta_v)$ that follows the specified concept while remaining fluent and coherent. This gives sentence-level control over hallucination types and broad coverage across taxonomy categories.

We then apply a post-hoc verifier defined as  $\textsc{Verify}_{\mathcal{M}}(s, \tilde{s})$, which queries an LLM to check whether $\tilde{s}$ is semantically equivalent to $s$, following the criteria of \citep{SemanticEquivalent_liang2025seca}:
{
\[
\small
\text{\textsc{Verify}}_{\mathcal{M}}(s, \tilde{s}) =
\begin{cases}
1, & \text{if $\tilde{s}$ is semantically equivalent to $s$} \\
0, & \text{otherwise,}
\end{cases}
\]
}
and discard any template with $\textsc{Verify}_{\mathcal{M}}(s, \tilde{s}) = 1$. Whereas the template-level check in Sec.~\ref{subsec:hallucination_template_construction} assesses \emph{applicability} before generation, this verifier operates on generated outputs for quality control. The complete pseudocode and prompt are provided in Appendix~\ref{app:algorithm} (Algorithm~\ref{alg:hallucination_injection_verified}) and Appendix~\ref{app:prompt_example}.

\section{Experimental Results}
\label{sec:eval}
\subsection{Task Formulation}
We define three sub-tasks that progressively evaluate a model’s ability to 
\emph{detect}, \emph{categorize}, and \emph{localize} hallucinated content in peer-review text.

\noindent\textbf{Task 1: Hallucination Detection.}
This task aims to determine whether hallucinated content is present, formulated as a binary classification problem at two levels of granularity: (1) \textit{Review-level}. Given a review $r$ with sentences $\mathcal{S}_r$, the model predicts whether the review contains hallucinated claims.(2) \textit{Sentence-level}. Given a sentence $s \in \mathcal{S}_r$, the model predicts whether the sentence contains hallucinated content.

\noindent\textbf{Task 2: Hallucination Type Classification.}
Given a hallucinated sentence $\tilde{s}$, together with supporting evidence from the corresponding source paper $p$, the model is required to perform multi-class classification and predict a single hallucination type from the coarse-grained label set $\mathcal{H} = \{ h_v \mid v \in \mathcal{V}^{(1)} \}$.

\noindent\textbf{Task 3: Hallucination Localization.}
To enable fine-grained error analysis, we formulate hallucination localization as a span identification task. Given a review $r$, the model is required to identify the spans corresponding to hallucinated content.

\subsection{Baselines}
\label{subsec:baselines}
We evaluate baselines under three paradigms for hallucination detection: (1) specialized verification frameworks, (2) prompting-based general-purpose LLMs, and (3) instruction-tuned LLMs. For hallucination type classification and localization, we evaluate prompting-based and instruction-tuned LLMs. Implementation details and prompt templates are provided in Appendix~\ref{app:baseline_implementation} and Appendix~\ref{app:prompt_example}.

\noindent\textbf{Specialized Verification Frameworks.} We evaluate four hallucination verifiers: \texttt{HHEM-2.1-Open}~\citep{Baseline_hhem-2.1-open} (consistency-based), \texttt{True-NLI}~\citep{Baseline_TrueNLI_laurer2022less} (entailment-based), a \texttt{\textsc{seNtLI}}-style retriever-verifier pipeline~\citep{Baseline_SENTLI_schuster2022stretching}, and \texttt{RefChecker}~\citep{Baseline_hu2024refchecker}.

\noindent\textbf{General-Purpose LLMs.}
We evaluate seven prompting-based LLMs, including \texttt{Qwen3-32B}, \texttt{Llama-3.3-70B}, \texttt{GPT-OSS-20B/120B}, \texttt{Mistral-Small-3.1-24B}, \texttt{RootSignals-Judge-Llama-70B}, and \texttt{GPT-5.2}. Following prior hallucination evaluation work~\citep{HalluDetect_HaluEval_li2023}, we additionally implement a Retrieval-Augmented LLM-as-a-Judge (\texttt{RA-LLM}) framework with three prompting strategies: \emph{Knowledge Retrieval (KR)}, \emph{Chain-of-Thought (CoT)}, and \emph{Sample Contrast (Contrast)}. For instruction-tuned baselines, we apply QLoRA-based 4-bit fine-tuning to \texttt{Qwen2.5-3B/7B-Instruct} and \texttt{Qwen3-32B}.

\noindent\textbf{Evidence Retrieval.}
For all specialized verifiers and \texttt{RA-LLM}, we apply BM25-based retrieval over the source paper to obtain a compact evidence context. For other prompting-based LLM baselines, we evaluate variants that consume the full paper content, along with LCS variants (detailed results are provided in Appendix~\ref{app:supp_experiment_result_ablation_RA-LLM}).

\subsection{Evaluation Metrics}
We evaluate models on the three tasks using following metrics. Details are deferred to Appendix~\ref{app:evaluation_metrics}.

\noindent\textbf{Hallucination Detection.}
We report \textit{Accuracy}, \textit{Precision}, \textit{Recall}, and \textit{F1}. Given the class imbalance in hallucination labels, we additionally report the \textit{Matthews Correlation Coefficient (MCC)}.

\noindent\textbf{Hallucination Type Classification.} 
We report \textit{Macro-F1}, which equally weights all hallucination categories, and \textit{Micro-F1}, which reflects overall instance-level performance.

\noindent\textbf{Hallucination Localization.}
We evaluate span identification using complementary token-level and span-level metrics. At the token level, we compute \textit{Token-F1} by converting predicted and gold spans into BIO tag sequences. At the span level, we report \textit{Exact Match Span-F1}, which requires exact boundary alignment between predicted and gold spans, and \textit{Overlap Span-F1}, which considers partially overlapping spans as correct predictions.

\subsection{Hallucination Detection (Task 1)}
\label{sec:exp_result-task1}
Tab.~\ref{tab:task1_detection} presents the performance of hallucination detection at both review and sentence levels on HalluPeer (NeurIPS 2024). Additional results on ICLR 2024 are provided in Appendix~\ref{app:supp_exp_in-domain_ICLR2024_task1}. Our observations are summarized below:

\noindent\textbf{Limitations of Specialized Verifications.}
Pre-trained verifiers (e.g., \texttt{HHEM-2.1-Open}, \texttt{True-NLI}, \texttt{\textsc{seNtLI}}, and \texttt{RefChecker}) generalize poorly to peer reviews. At the review level, all models exhibit near-random performance with MCC scores $\leq 0.03$. While \texttt{HHEM-2.1-Open} performs slightly better at the sentence level—likely due to fine-tuning on RAG hallucination datasets—its overall effectiveness remains limited by the domain gap between general RAG settings and scientific peer reviews. Overall, these findings suggest that existing hallucination verifiers, whether consistency- or entailment-based, fail to transfer to scientific review verification. Verifying technical review claims against full-length papers requires complex multi-hop reasoning that differs substantially from the open-domain RAG, NLI, and general factual consistency datasets used to train these models.

\noindent\textbf{Impact of Prompting Strategies.}
Among the RA-LLM variants, prompting strategies exhibit distinct performance. At the sentence level, \texttt{RA-LLM (KR)}, which augments prompts with few-shot demonstrations, achieves the strongest performance (MCC 0.61, Accuracy 0.82), outperforming both explicit reasoning (\texttt{RA-LLM (CoT)}) and contrastive prompting (\texttt{RA-LLM (Contrast)}). One explanation is that \texttt{CoT} and \texttt{Contrast} prompting rely more heavily on the completeness of source paper evidence. Since evidence is first filtered through BM25 retrieval, the retrieved context may not provide sufficient information for multi-step reasoning or contradiction analysis. In this setting, concise few-shot demonstrations offer a more stable supervision signal, allowing \texttt{RA-LLM (KR)} to remain effective. 

Under standard prompting, frontier LLMs such as \texttt{Qwen3-32B}, \texttt{GPT-OSS-120B}, and \texttt{GPT-5.2} achieve competitive sentence-level performance, with F1 and MCC scores comparable to RA-LLM variants. However, they consistently underperform at the review level, indicating difficulty in maintaining globally consistent verification over long-form contexts and aggregating evidence coherently.

\noindent\textbf{The Superiority of Domain-Specific Fine-tuning.}
We observe substantial gains from domain-specific instruction tuning. Fine-tuned models consistently dominate both review- and sentence-level evaluation. Remarkably, even the relatively compact \texttt{Qwen2.5-3B} significantly outperforms all prompting-based zero-shot methods, including frontier models such as \texttt{GPT-5.2} and \texttt{Llama-3.3-70B}. The scaled-up \texttt{Qwen3-32B} (Fine-tuned) achieves the strongest overall performance, reaching F1 scores of 0.90 and 0.91 at the review and sentence levels, respectively, together with a sentence-level MCC of 0.87. These findings suggest that peer-review hallucination detection depends heavily on domain-specific verification patterns that are not sufficiently captured by generic zero-shot prompting alone.

\begin{table}[ht] 
\centering
\setlength{\tabcolsep}{3pt}
\fontsize{6.05}{11}\selectfont
\caption{\textbf{Task 1 results on HalluPeer (NeurIPS 2024).} Review-/sentence-level results are shown before/after the slash. Results are reported under the default evidence and prompting settings described in Sec.~\ref{subsec:baselines}.} 
\label{tab:task1_detection}
\begin{tabularx}{0.5\textwidth}{Xccccc}
\toprule
\textbf{Specialized Verification} & Acc. & Prec. & Rec. & F1 &  MCC \\
\cmidrule(lr){1-1}\cmidrule(lr){2-6}
\texttt{HHEM-2.1-Open} & 0.51 / \textbf{0.64} & 0.51 / \textbf{0.47} & 0.55 / \textbf{0.63} & 0.53 / \textbf{0.54} & 0.02 / \textbf{0.26} \\
\texttt{True-NLI} & \textbf{0.52} / 0.54 & \textbf{0.52} / 0.37 & \textbf{0.56} / 0.52 & \textbf{0.54} / 0.43 & \textbf{0.03} / 0.07 \\
\texttt{\textsc{seNtLI}} & 0.51 / 0.61 & 0.51 / 0.44 & 0.54 / 0.55 & 0.52 / 0.49 & 0.02 / 0.18 \\
\texttt{Refchecker} & 0.51 / 0.49 & 0.51 / 0.34 & 0.45 / 0.51 & 0.48 / 0.40 & \textbf{0.03} / -0.01 \\
\midrule
\midrule
\textbf{LLM (Prompting)} & Acc. & Prec. & Rec. & F1 &  MCC \\
\cmidrule(lr){1-1}\cmidrule(lr){2-6}
\texttt{RA-LLM (KR)} & 0.61 / \textbf{0.82} & 0.68 / \textbf{0.74} & 0.41 / 0.74 & 0.51 / \textbf{0.74} & 0.24 / \textbf{0.61} \\
\texttt{RA-LLM (CoT)} & 0.61 / 0.77 & 0.63 / 0.63 & 0.52 / 0.78 & 0.57 / 0.70 & 0.22 / 0.53 \\
\texttt{RA-LLM (Contrast)} & 0.57 / 0.73 & 0.55 / 0.57 & 0.77 / \textbf{0.84} & 0.64 / 0.68 & 0.15 / 0.49 \\
\texttt{Qwen3-32B} & \textbf{0.63} / 0.79 & 0.63 / 0.66 & 0.61 / 0.74 & 0.62 / 0.70 & \textbf{0.25} / 0.54 \\
\texttt{Llama-3.3-70B} & 0.57 / 0.70 & \textbf{0.74} / 0.55 & 0.22 / 0.63 & 0.34 / 0.59 & 0.20 / 0.36 \\
\texttt{Mistral-Small-3.1} & 0.61 / 0.73 & 0.63 / 0.58 & 0.52 / 0.74 & 0.57 / 0.65 & 0.22 / 0.45 \\
\texttt{GPT-OSS-20B} & 0.59 / 0.78 & 0.56 / 0.65 & 0.79 / 0.76 & 0.66 / 0.70 & 0.19 / 0.53 \\
\texttt{GPT-OSS-120B} & 0.56 / 0.78 & 0.53 / 0.64 & \textbf{0.92} / 0.81 & 0.67 / 0.72 & 0.17 / 0.55 \\
\texttt{Judge-Llama-70B} & 0.57 / 0.70 & \textbf{0.74} / 0.55 & 0.23 / 0.64 & 0.35 / 0.59 & 0.20 / 0.36 \\
\texttt{GPT-5.2} & 0.58 / 0.80 & 0.55 / 0.70 & 0.89 / 0.73 & \textbf{0.68} / 0.71 & 0.21 / 0.56 \\
\midrule
\midrule
\textbf{LLM (Fine-tuned)} & Acc. & Prec. & Rec. & F1 &  MCC \\
\cmidrule(lr){1-1}\cmidrule(lr){2-6}
\texttt{Qwen2.5-3B} & 0.84 / 0.85 & 0.86 / 0.71 & 0.83 / \textbf{0.93} & 0.84 / 0.81 & 0.69 / 0.70 \\
\texttt{Qwen2.5-7B} & 0.87 / 0.91 & 0.86 / 0.82 & \textbf{0.89} / \textbf{0.93} & 0.87 / 0.87 & 0.74 / 0.80 \\
\texttt{Qwen3-32B} & \textbf{0.90} / \textbf{0.94} & \textbf{0.96} / \textbf{0.94} & 0.85 / 0.89 & \textbf{0.90} / \textbf{0.91} & \textbf{0.81} / \textbf{0.87} \\
\bottomrule
\end{tabularx}
\end{table}


\subsection{Hallucination Type Classification (Task 2)}
\label{sec:exp_result-task2}
\begin{figure}
    \centering
    \includegraphics[width=\columnwidth]{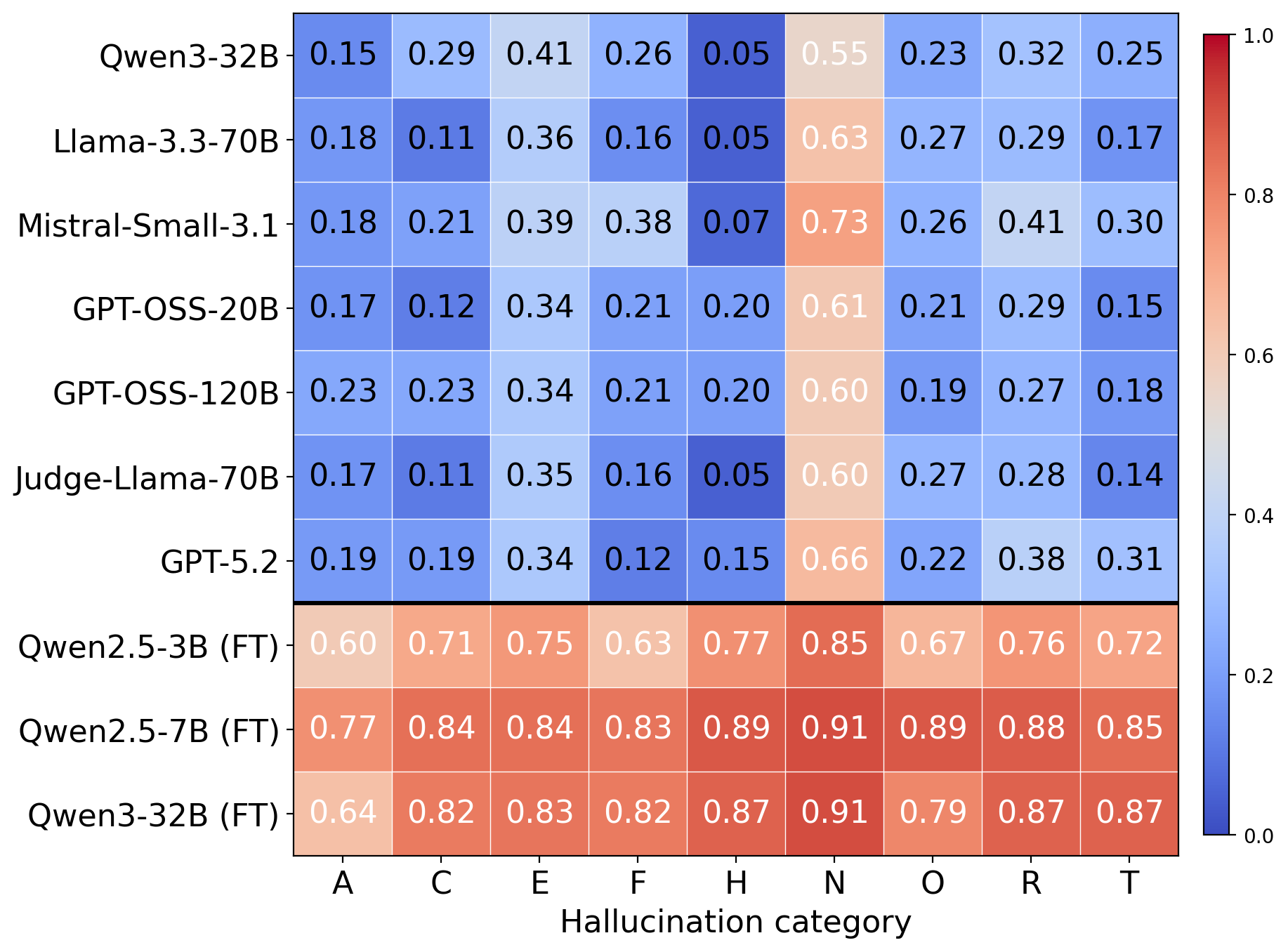}
    \caption{\textbf{Visualization of Task 2 sentence-level per-label F1 results.} Columns denote hallucination categories: A (\textit{Attribution Failure}), C (\textit{Context-based Meaning Error}), E (\textit{Entity}), F (\textit{False Concatenation}), H (\textit{Hyperbole}), N (\textit{Number}), O (\textit{Overgeneralization}), R (\textit{Reasoning Error}), and T (\textit{Temporal}).
    \label{fig:task2_per_label_heatmap}}
\end{figure}

Tab.~\ref{tab:task2_classification_overall} presents the overall results of hallucination type classification on HalluPeer (NeurIPS 2024), while Fig.~\ref{fig:task2_per_label_heatmap} visualizes per-label performance. Additional results on ICLR 2024 are provided in Appendix~\ref{app:supp_exp_in-domain_ICLR2024_task2}. Our key findings include:

\noindent\textbf{Limitations of Zero-shot Prompting.}
All prompting-based LLMs perform poorly, especially at the review level, where the best Macro-F1 reaches only 0.19. Interestingly, larger models do not consistently yield better performance. For example, the smaller \texttt{Mistral-Small-3.1} outperforms larger models such as \texttt{Llama-3.3-70B} and \texttt{GPT-OSS-120B} on both Macro-F1 and Micro-F1. These findings suggest that hallucination category recognition relies more on robust scientific verification behavior than model scale.

\noindent\textbf{Review-level Categorization Remains Challenging.}
A clear gap exists between sentence- and review-level classification across all prompting-based LLMs, with review-level performance consistently weak. One possible explanation is that hallucination evidence in peer reviews is highly localized, with only a small hallucinated span embedded within largely correct content. Under review-level classification, such sparse error signals may be diluted by surrounding context or overlooked due to long-context reasoning limitations such as lost-in-the-middle effects, making review-level categorization substantially more difficult.

\noindent\textbf{Semantic Difficulty Varies Across Hallucination Types.}
Under prompting-only settings, performance varies substantially across hallucination categories. Categories with explicit lexical cues, such as \textit{Entity} and \textit{Number}, achieve relatively higher F1 scores. In contrast, semantically complex categories requiring contextual grounding or multi-hop reasoning, including \textit{Context-based Meaning Error}, \textit{Hyperbole}, and \textit{Temporal}, remain highly challenging, with several models collapsing to near-zero review-level F1. These findings suggest that current LLMs are more effective at detecting surface-level factual inconsistencies than deeper semantic distortions in scientific peer reviews.

\noindent\textbf{Domain-specific Fine-tuning Enables Robust Error Taxonomy Recognition.}
Fine-tuning closes the gaps identified above. Fine-tuned models substantially outperform their zero-shot counterparts, particularly on semantically challenging categories. For example, \textit{Hyperbole} improves from near zero to 0.72/0.87, \textit{Temporal} to 0.48/0.87, and \textit{Context-based Meaning Error} to 0.56/0.82. These results suggest that the semantic difficulties observed under zero-shot prompting can be substantially mitigated through domain-specific adaptation.


\begin{table}[ht] 
\centering
\setlength{\tabcolsep}{10pt}
\fontsize{8}{11}\selectfont
\caption{\textbf{Task 2 overall results on HalluPeer (NeurIPS 2024).} Review-/sentence-level results are shown before/after the slash. Detailed per-label F1 results are provided in Appendix~\ref{app:task2_per_label}.}
\label{tab:task2_classification_overall}
\begin{tabularx}{\columnwidth}{Xcc}
\toprule
\textbf{LLM (Prompting)} & Macro-F1 & Micro-F1 \\
\cmidrule(lr){1-1}\cmidrule(lr){2-3}
\texttt{Qwen3-32B} & 0.15 / 0.28 & 0.17 / 0.30 \\
\texttt{Llama-3.3-70B} & 0.14 / 0.24 & 0.17 / 0.27 \\
\texttt{Mistral-Small-3.1} & \textbf{0.19} / \textbf{0.33} & \textbf{0.22} / \textbf{0.35} \\
\texttt{GPT-OSS-20B} & 0.13 / 0.25 & 0.15 / 0.28 \\
\texttt{GPT-OSS-120B} & 0.16 / 0.25 & 0.18 / 0.26 \\
\texttt{Judge-Llama-70B} & 0.14 / 0.23 & 0.17 / 0.26 \\
\texttt{GPT-5.2} & 0.14 / 0.28 & 0.17 / 0.31 \\
\midrule
\midrule
\textbf{LLM (Fine-tuned)} & Macro-F1 & Micro-F1 \\
\cmidrule(lr){1-1}\cmidrule(lr){2-3}
\texttt{Qwen2.5-3B} & 0.49 / 0.72 & 0.52 / 0.72 \\
\texttt{Qwen2.5-7B} & 0.53 / \textbf{0.86} & 0.55 / \textbf{0.86} \\
\texttt{Qwen3-32B} & \textbf{0.59} / 0.82 & \textbf{0.59} / 0.84 \\
\bottomrule
\end{tabularx}
\end{table}


\subsection{Hallucination Localization (Task 3)}
\label{sec:exp_result-task3}
Tab.~\ref{tab:task3_localization} presents the results for hallucination span localization on HalluPeer (NeurIPS 2024). Additional results on ICLR 2024 are provided in Appendix~\ref{app:supp_exp_in-domain_ICLR2024_task3}. Key findings include:

\noindent\textbf{Limitations of Zero-shot Prompting.}
All prompting-based models exhibit limited span localization capability, particularly under strict boundary matching metrics. Among zero-shot methods, \texttt{GPT-5.2} achieves the strongest performance, reaching 0.58 Token-F1 and 0.46 Exact Span-F1. However, overall performance remains moderate even for frontier models, suggesting that hallucination localization in peer reviews is inherently challenging. Similar to review-level hallucination categorization, hallucinated content is often sparse and embedded within otherwise correct scientific critique, making precise grounding and boundary identification considerably more difficult.

\noindent\textbf{Challenge of Exact Boundary Detection.}
A consistent gap is observed between Overlap Span-F1 and Exact Span-F1 across models. For example, \texttt{GPT-5.2} achieves 0.58 Overlap Span-F1 but only 0.46 Exact Span-F1, indicating that models can often localize hallucinated regions approximately but struggle to determine precise token boundaries. This issue is particularly pronounced in peer reviews, where hallucinations appear as localized semantic distortions embedded within otherwise coherent scientific arguments.

\noindent\textbf{Fine-tuning Dramatically Improves Hallucination Grounding.}
Fine-tuned models substantially outperform prompting-based approaches across all metrics. Notably, \texttt{Qwen3-32B} achieves 0.91 Token-F1 and 0.86 Exact Span-F1. The gains suggest that accurate hallucination localization requires specialized supervision for token-level grounding and error boundary identification, which cannot be reliably induced through only zero-shot prompting.

\begin{table}[ht] 
\centering
\setlength{\tabcolsep}{4pt}
\fontsize{8}{11}\selectfont
\caption{\textbf{Task 3 results on HalluPeer (NeurIPS 2024).} Review-level results are reported under default evidence retrieval and prompting settings described in Sec.~\ref{subsec:baselines}.}
\label{tab:task3_localization}
\begin{tabularx}{0.5\textwidth}{Xccc}
\toprule
\textbf{LLM (Prompting)} & Token-F1 & Exact Span-F1 & Overlap Span-F1 \\
\cmidrule(lr){1-1}\cmidrule(lr){2-4}
\texttt{Qwen3-32B} & 0.46 & 0.30 & 0.50 \\
\texttt{Llama-3.3-70B} & 0.49 & 0.35 & 0.49 \\
\texttt{Mistral-Small-3.1} & 0.40 & 0.27 & 0.40 \\
\texttt{GPT-OSS-20B} & 0.51 & 0.38 & 0.49 \\
\texttt{GPT-OSS-120B} & 0.55 & 0.41 & 0.52 \\
\texttt{Judge-Llama-70B} & 0.50 & 0.36 & 0.49 \\
\texttt{GPT-5.2} & \textbf{0.58} & \textbf{0.46} & \textbf{0.58} \\
\midrule
\midrule
\textbf{LLM (Fine-tuned)} & Token-F1 & Exact Span-F1 & Overlap Span-F1 \\
\cmidrule(lr){1-1}\cmidrule(lr){2-4}
\texttt{Qwen2.5-3B} & 0.85 & 0.79 & 0.82 \\
\texttt{Qwen2.5-7B} & 0.84 & 0.82 & 0.84 \\
\texttt{Qwen3-32B} & \textbf{0.91} & \textbf{0.86} & \textbf{0.90} \\
\bottomrule
\end{tabularx}
\end{table}



\subsection{Cross-Venue Transferability}
\label{sec:cross-venue}
To evaluate the cross-venue transferability of our fine-tuned detectors, we fine-tune each on one venue split and evaluate it directly on the completely held-out venue, considering both transfer directions. Results are reported in Tab.~\ref{tab:cross_venue_task1}--\ref{tab:cross_venue_task3}.

\noindent\textbf{Task 1 (Detection).}
Domain-specific fine-tuning demonstrates robust generalization across venues.
As shown in Tab.~\ref{tab:cross_venue_task1}, Qwen3-32B achieves review-/sentence-level F1 scores of 0.90/0.91 when trained on NeurIPS 2024 and tested on ICLR 2024, and 0.86/0.93 in the reverse direction. These results indicate that the learned detection capability transfers across different conference distributions.

\noindent\textbf{Task 2 (Type Classification).}
The performance gap between review- and sentence-level results persists across venues.
As shown in Tab.~\ref{tab:cross_venue_task2_overall}, Qwen3-32B achieves sentence-level Micro-F1 scores of 0.83 and 0.86 for NeurIPS~$\rightarrow$~ICLR and ICLR~$\rightarrow$~NeurIPS, respectively, indicating that the injected hallucination types remain recognizable across different conference distributions.

\noindent\textbf{Task 3 (Localization).} Cross-venue span localization remains highly accurate. Tab.~\ref{tab:cross_venue_task3} shows that the fine-tuned Qwen3-32B model preserves excellent boundary grounding on unseen venues, reaching 0.91 Token-F1 and 0.88 Exact Span-F1 when transferring from NeurIPS to ICLR. The performance in the ICLR $\rightarrow$ NeurIPS direction is comparable (0.92 Token-F1 and 0.87 Exact Span-F1), confirming the general applicability and robustness of our localization training.

\noindent\textbf{Comparison with In-Domain Results.}
We further compare cross-venue transfer with the corresponding in-domain results on the same test venue. 
For NeurIPS~$\rightarrow$~ICLR, Qwen3-32B achieves Task 1 F1 of 0.90/0.91 and Task 3 Token-F1 of 0.91, compared with 0.89/0.94 and 0.93, respectively, for the ICLR in-domain results (Appendix~\ref{app:supp_exp_in-domain_ICLR2024_task1}, \ref{app:supp_exp_in-domain_ICLR2024_task3}).
For ICLR $\rightarrow$ NeurIPS, it achieves Task~1 F1 of 0.86/0.93 and Task~3 Token-F1 of 0.92, compared with the NeurIPS in-domain results of 0.90/0.91 and 0.91, respectively (Tab.~\ref{tab:task1_detection}, \ref{tab:task3_localization}).
The relatively small performance differences across transfer directions suggest that the detectors do not rely strongly on venue-specific writing styles or formatting, but instead internalize the structural definitions of peer-review hallucinations.

\subsection{Cross-Generation Ablation}
\label{sec:cross_generation}
To examine whether our fine-tuned detectors rely on generator-specific artifacts, we construct additional test sets using Mistral-Small-3.1 and Llama-3.3-70B as hallucination injectors and verifiers, while keeping the taxonomy and construction protocol unchanged. Detectors are trained exclusively on Qwen3-32B-injected data and evaluated on these unseen-generator test sets. \textbf{Across all three tasks, performance remains largely stable under generator changes.} For Task~1, the F1 shift is only 0.01--0.02 on average; Task~2 shows shifts generally within 0.05, while Task~3 Token-F1 remains within 0.02 of the in-domain results. Detailed results and per-task analyses are provided in Appendix~\ref{app:cross_generation}.

\subsection{Evaluation on Authentic Reviews}
To quantitatively assess whether detectors trained on synthetic hallucinations transfer to authentic reviewer errors, we construct a manually annotated set of 1,161 independent NeurIPS 2024 reviews, identifying 20 genuine reviewer hallucinations. The fine-tuned detectors are trained exclusively on synthetic HalluPeer ICLR 2024 data and have no access to these authentic annotations. As reported in Appendix~\ref{app:real_review_quantitative}, Qwen3-32B recovers all 20 authentic hallucinations (TPR $=100.0\%$) at FPR $=22.1\%$, substantially improving recall over zero-shot baselines. These results provide quantitative evidence that detectors trained on synthetic hallucinations can transfer to naturally occurring reviewer errors.

\section{Alignment with Real Review Errors}
\label{sec:sim_to_real}
\noindent\textbf{Hallucination Types in Authentic Reviews}
To evaluate whether the hallucination patterns defined by our taxonomy occur in authentic reviews, we conduct a case study on 13,803 human-written NeurIPS 2024 reviews. Our fine-tuned detector flags potential hallucinations, and we manually inspect 200 flagged instances, checking whether each constitutes a genuine hallucination and whether the predicted type matches. As shown in Tab.~\ref{tab:hallu-taxonomy-examples}--\ref{tab:hallu-taxonomy-examples-3}, the identified review errors align with the hallucination patterns defined by our taxonomy. These findings provide evidence that the hallucination patterns defined by HalluPeer correspond to errors occurring in authentic peer reviews. The annotation protocol details are provided in Appendix~\ref{app:real_review_annotation}.

\noindent\textbf{False Positives from Paper Claim Quotation.}
Our manual inspection reveals a source of false positives arising from reviewer quotations of exaggerated or unsupported claims in the submitted paper. Consequently, review texts may contain hallucination-like statements that originate from the paper rather than the reviewer. This highlights a key challenge for peer-review verification systems: distinguishing reviewer-generated hallucinations from the propagation of unsupported claims in the submitted paper. Such source-aware attribution is important for distinguishing genuine reviewer errors from valid criticism.

\section{Conclusion}
We present \textbf{HalluPeer}, a taxonomy-driven benchmark for hallucination detection in peer reviews. HalluPeer formulates review auditing as a \emph{paper-grounded verification} problem requiring long-context reasoning over manuscripts. To support systematic evaluation, we proposed a hierarchical taxonomy and an aspect-aware injection pipeline for generating realistic hallucinated reviews.

Experiments show that existing verifiers struggle to distinguish unsupported claims from legitimate scientific critique, while domain-specific fine-tuning substantially improves performance. Our evaluations on authentic reviews provide evidence that HalluPeer-defined hallucination patterns occur in real peer reviews, while source-aware attribution remains a key challenge for trustworthy AI-assisted peer review.


\section*{Acknowledgments} 
This work is partially supported by the National Science and Technology Council, Taiwan, under Grant: NSTC-115-2923-E-A49 -010-MY5.

\section{Limitations}
We acknowledge several limitations in our work.
\noindent\textbf{Coverage of naturally occurring errors.} 
Naturally occurring review hallucinations are sparse and require domain expertise to identify, making large-scale real-world annotation prohibitively labor-intensive. Rather than modeling an unobservable real-world distribution, HalluPeer provides a controlled, practically motivated framework that covers plausible error patterns, supporting future research on automated peer-review auditing and review quality tracking.

\noindent\textbf{Synthetic nature of the dataset.} While our injection pipeline is designed to simulate realistic errors via aspect-conditioning and style preservation, the resulting dataset remains synthetic. The distribution of injected hallucinations may not perfectly reflect the subtle, drift-based errors found in reviews in real-world scenarios. Naturally occurring hallucinations might involve more complex reasoning failures that are difficult to simulate through localized editing.

\noindent\textbf{Domain specificity.} Our data source is restricted to computer science conferences hosted on OpenReview, primarily due to the scarcity of publicly available peer-review datasets in other fields. Reviewing norms, claim structures, and evidence densities vary significantly across scientific disciplines. Consequently, the taxonomy and detection models developed on HalluPeer may not generalize zero-shot to other domains without specific adaptation.

\noindent\textbf{Scope of hallucination definition.} We restrict our definition of hallucination to factual inconsistency with respect to the submission content. We do not address subjective aspects of the review process, such as unfair novelty judgments, tonal issues, or the validity of critiques regarding potential future work. These subjective elements are critical for high-quality peer review but require different evaluation frameworks beyond factual grounding.

\noindent\textbf{Model-induced bias.} Since the taxonomy is derived via LLM-guided decomposition, the resulting structure may inherit the inductive biases of the proposer models. These biases could influence how hallucination types are grouped or defined. Consequently, the constructed hierarchy represents a model-centric perspective on error categorization rather than a canonical or exhaustive standard.

\noindent\textbf{Selection bias from meta-review alignment.} Our use of meta-review alignment as a selection heuristic may bias HalluPeer toward reviews whose main points were reflected in the final assessment, rather than the full distribution of review quality. This is a deliberate trade-off in controlled benchmark construction: a more selective base-review set improves label reliability and reduces the likelihood of selecting unreliable reviews, while broader sampling would provide greater distributional coverage at the cost of noisier supervision.


\section{Ethical Considerations}
While this work aims to enhance the integrity of the academic peer-review process, we acknowledge the following ethical implications of using scholarly data.

\noindent\textbf{Dual-Use Risks.} Our hallucination injection pipeline poses potential dual-use risks: although developed for benchmarking detection systems, it could be misused to generate more convincing hallucinated reviews. To mitigate this concern, we restrict our taxonomy and generation scripts to defensive research purposes.

\noindent\textbf{Integrity of the Peer-Review Process.} 
Our research synthetically corrupts human-written reviews to construct negative samples. No hallucinated reviews were submitted to real venues or used in editorial decisions. The dataset is intended solely for offline training and evaluation. We advocate human-in-the-loop review systems, where AI functions as a diagnostic aid rather than an autonomous decision-maker.

\bibliography{custom}

\appendix


\section{Statistics}
\subsection{HalluPeer Dataset Statistic}
\label{app:data_stats}
We collect peer-review data from OpenReview, covering two major machine learning conferences: ICLR (2019--2024) and NeurIPS (2021--2024). Tab.~\ref{tab:dataset_statistics} summarizes the statistics of HalluPeer across venues and years.

\noindent\textbf{Data Scale.} To ensure balanced representation, we uniformly sample 1,200 papers from each venue-year pair. The resulting dataset contains 12,000 papers, 38,063 reviews, and over 1.02M review sentences, providing substantial coverage for hallucination detection and analysis.

\noindent\textbf{Hallucination Distribution.} Our injection pipeline constructs more than 10.1M hallucination templates in total. The dataset maintains a consistent injection density with an average of 9.86 templates per sentence across different venue-year pairs. Furthermore, the number of injected hallucinations can be flexibly adjusted, enabling controllable construction of positive and negative instances for diverse training and evaluation settings.

\begin{table*}[h]
\centering
\caption{\textbf{Statistics of the HalluPeer Dataset.} We report the total number of source papers, reviews, parsed review sentences, and hallucination templates (\textit{Tpls}) for each venue and year.}
\label{tab:dataset_statistics}
\setlength{\tabcolsep}{8pt}
\begin{tabular}{l c c c c c c}
\toprule
\textbf{Venue} & \textbf{Year} & \textbf{\# Papers} & \textbf{\# Reviews} & \textbf{\# Rev. Sents} & \textbf{\# Hallu. Tpls} & \textbf{Avg Tpls/Sent} \\
\midrule
\multirow{4}{*}{NeurIPS} 
& 2021 & 1,200 & 3,955 & 110,888 & 1,143,027 & 10.31 \\
& 2022 & 1,200 & 3,540 & 99,935  & 965,645   & 9.66  \\
& 2023 & 1,200 & 4,311 & 118,108 & 1,068,969 & 9.05  \\
& 2024 & 1,200 & 3,923 & 105,118 & 865,805   & 8.24  \\
\midrule
\multirow{6}{*}{ICLR} 
& 2019 & 1,200 & 3,143 & 71,031  & 800,004   & 11.26 \\
& 2020 & 1,200 & 2,970 & 69,032  & 760,750   & 11.02 \\
& 2021 & 1,200 & 3,935 & 105,287 & 1,115,234 & 10.59 \\
& 2022 & 1,200 & 4,038 & 119,669 & 1,262,116 & 10.55 \\
& 2023 & 1,200 & 4,081 & 118,794 & 1,151,737 & 9.70  \\
& 2024 & 1,200 & 4,167 & 110,989 & 1,015,813 & 9.15  \\
\midrule
\textbf{Total / Avg.} & \textbf{--} & \textbf{12,000} & \textbf{38,063} & \textbf{1,028,851} & \textbf{10,149,100} & \textbf{9.86} \\
\bottomrule
\end{tabular}
\end{table*}

\subsection{HalluPeer Taxonomy Statistics}
\label{app:taxonomy_stats}
We analyze the structural properties of our constructed peer-review hallucination taxonomy. The taxonomy is organized as a hierarchical tree structure with a maximum depth of 3. Tab.~\ref{tab:taxonomy_stats} summarizes the distribution of nodes and branching factors across different depths.

\noindent\textbf{Tree dimensions.} The complete taxonomy comprises a total of 265 nodes. Among these, there are 205 distinct leaf nodes which represent the terminal, fine-grained hallucination types used for injection. The hierarchy expands from the root to a maximum depth of 3, ensuring a granular decomposition of review-specific errors.

\noindent\textbf{Depth distribution.} The node distribution demonstrates a comprehensive refinement process. The operational definitions are located at the deepest level. Specifically, 200 out of the 205 leaf nodes (approximately 98\%) reside at Depth 3. This bottom-heavy structure indicates that the recursive decomposition successfully transforms abstract high-level concepts into specific, atomic instructions suitable for localized injection.

\noindent\textbf{Branching characteristics.} The branching factor analysis illustrates the expansion rate of the semantic space. The root node is initialized into 9 coarse categories (Depth 1). The average branching factor then transitions from 6.11 at Depth 1 to 4.00 at Depth 2. This suggests a consistent expansion strategy where broad categories are broken down into approximately 4 to 6 subtypes at each intermediate step, balancing breadth and depth before reaching the terminal leaf nodes.

\subsection{HalluPeer Taxonomy Examples}

\noindent\textbf{Depth 1: Coarse-grained Hallucinations.} 
Following prior work on hallucination categorization \citep{HalluDetect_HalluMeasure_akbar2024}, we adopt nine broad categories that capture common forms of unsupported generation: \textit{Number}, \textit{Entity}, \textit{False Concatenation}, \textit{Attribution Failure}, \textit{Overgeneralization}, \textit{Reasoning Error}, \textit{Hyperbole}, \textit{Temporal}, and \textit{Context-based Meaning Error}. The definitions for these categories are provided in Tab.~\ref{tab:hallucination-typology}.

\noindent\textbf{Depth 2: Domain-Specific Hallucinations.} 
Our goal at this level is to expand each coarse concept into a set of review-specific subtypes that are (i) mutually distinguishable, (ii) consistent in granularity across siblings, and (iii) sufficiently operational to facilitate both dataset annotation and controllable hallucination injection. The prompt template used for this expansion is detailed in Fig.~\ref{fig:prompt_hallucination_taxonomy_generation}. As demonstrated in Fig.~\ref{fig:taxonomy_examples}, a coarse-grained concept like ``Number'' is decomposed into academic-contextualized categories such as ``Year Discrepancy'' and ``Statistical Value Mismatch.''

\noindent\textbf{Depth 3: Fine-grained Hallucinations.} 
As further illustrated in Fig.~\ref{fig:taxonomy_examples}, we decompose the domain-specific subtypes into precise, operational leaf nodes. For instance, ``Year Discrepancy'' is instantiated into highly specific error instructions such as ``Event Year Fabrication'' and ``Citation Year Substitution.''

\definecolor{headbar}{HTML}{ECEFF1}

\begin{table*}[th]
  \centering
  \footnotesize
  \setlength{\tabcolsep}{4pt}
  \renewcommand{\arraystretch}{1.25}
    \caption{Typology of hallucination categories with their descriptions.}
  \label{tab:hallucination-typology}
  \begin{tabularx}{\textwidth}{@{}l >{\raggedright\arraybackslash}X@{}}
    \toprule
    \textbf{Category} & \textbf{Description} \\
    \midrule
    \hnumber       & A claim has a different number than the original context (e.g.\ 20\% vs.\ 0.7\%). Any number, including year, dimensions, ages, etc. \\
    \addlinespace[2pt]
    \hentity       & A claim includes swapped, incorrectly specified, or inserted noun phrases (e.g.\ one named entity used in a context where another word is expected). \\
    \addlinespace[2pt]
    \hfalseconcat  & A claim incorrectly combines information about multiple entities or events. \\
    \addlinespace[2pt]
    \hattribution  & A claim lacks proper attribution, either crediting the wrong source or presenting information as fact without citation. \\
    \addlinespace[2pt]
    \hovergen      & A claim is based on accurate contextual information but is too broad or too general to be supported by the context. \\
    \addlinespace[2pt]
    \hreasoning    & A claim is based on accurate contextual information but contains a reasoning error or makes an unsupported conclusion. \\
    \addlinespace[2pt]
    \hhyperbole    & A claim is based on accurate information but exaggerated or overstated. \\
    \addlinespace[2pt]
    \htemporal     & A claim does not accurately incorporate tense, modality (e.g.\ \emph{might} vs.\ \emph{will}), or time reference in relation to the context. \\
    \addlinespace[2pt]
    \hcontext      & A claim includes incorrect interpretation of idiomatic language, homonyms, or words with multiple meanings, therefore failing to capture the intended meaning. \\
    \bottomrule
  \end{tabularx}
\end{table*}

\begin{figure*}[ht]
\ttfamily
\footnotesize
\setlength{\DTbaselineskip}{13pt} 
\dirtree{%
.2 \textcolor{depthOne}{Number (Depth: 1)}.
.3 \textcolor{descColor}{A claim has a different number than the original context (e.g. 20\% vs. 0.7\%). Any number, including year, dimensions, ages, etc.}.
.3 \textcolor{depthTwo}{Year Discrepancy (Depth: 2)}.
.4 \textcolor{descColor}{\emph{The review text cites a publication or event year that differs from the original context.}}.
.4 \textcolor{depthThree}{Event Year Fabrication (Depth: 3)}.
.5 \textcolor{descColor}{\emph{The review text references a scientific event, discovery, or milestone as occurring in a year that differs from the actual year in the original context.}}.
.4 \textcolor{depthThree}{Citation Year Substitution (Depth: 3)}.
.5 \textcolor{descColor}{\emph{The review text cites a source with a year that is different from the year listed in the original context for that source.}}.
.4 \textcolor{depthThree}{Grant or Funding Year Mismatch (Depth: 3)}.
.5 \textcolor{descColor}{\emph{The review text references a grant, funding award, or institutional support as being issued or active in a year that differs from the original context.}}.
.3 \textcolor{depthTwo}{Statistical Value Mismatch (Depth: 2)}.
.4 \textcolor{descColor}{\emph{The review text reports a statistical value (e.g., p-value, mean, standard deviation) that differs from the original context.}}.
.4 \textcolor{depthThree}{P-Value Mismatch (Depth: 3)}.
}
\medskip 
\caption{\textbf{Example of the recursive taxonomy decomposition.} This figure illustrates a subset of our peer-review hallucination taxonomy. A general hallucination concept (e.g., ``Number'') is recursively decomposed into intermediate subcategories (e.g., ``Year Discrepancy'') and ultimately into fine-grained leaf nodes (e.g., ``Citation Year Substitution''). These operational leaf nodes provide concrete instructions that enable highly controllable LLM-based hallucination injection.}
\label{fig:taxonomy_examples}
\end{figure*}


\begin{table}[h]
\centering
\fontsize{8}{11}\selectfont
\caption{\textbf{Structural Statistics of the Hallucination Taxonomy.} We report the total nodes, leaf nodes, and average branching factor. The high count at Depth 3 highlights the fine-grained nature of our taxonomy.}
\label{tab:taxonomy_stats}
\begin{tabularx}{0.5\textwidth}{crrr}
\toprule
\textbf{Depth} & \textbf{Total Nodes} & \textbf{Leaf Nodes} & \textbf{Avg. Branch. Factor} \\
\midrule
0 & 1 & 0 & 9.00 \\
1 & 9 & 0 & 6.11 \\
2 & 55 & 5 & 4.00 \\
3 & 200 & 200 & 0.00 \\
\midrule
\textbf{Total} & \textbf{265} & \textbf{205} & \textbf{--} \\
\bottomrule
\end{tabularx}
\end{table}

\section{Implementation Details}
\label{app:implementation_details}
\subsection{Model Configuration}
\label{app:model_config}
Unless otherwise specified, we employ \texttt{Qwen3-32B} as the backbone LLM for most components in the HalluPeer dataset construction pipeline. This includes the human-written review filtering ($\textsc{Filter}_{\mathcal{M}}$), template feasibility checking ($\textsc{Check}_{\mathcal{M}}$), hallucination template injection ($\textsc{Inject}_{\mathcal{M}}$), and post-hoc semantic verification ($\textsc{Verify}_{\mathcal{M}}$) modules.

Taxonomy generation and refinement follow the multi-model ensemble procedure described in Appendix~\ref{app:ensemble_details}. In particular, recursive taxonomy decomposition ($\textsc{Expand}_{\mathcal{M}}$) and hallucination concept description generation ($\textsc{Describe}_{\mathcal{M}}$) are independently performed using \texttt{Qwen3-32B}, \texttt{Llama-3.3-70B}, and \texttt{Mistral-Small-3.1-24B}.

To minimize variance and ensure reproducibility, we set the temperature to 0 for all generation operations and explicitly disable extended reasoning or ``thinking'' modes. The maximum generation length is adjusted according to the requirements of each module.

\subsection{Review Aspect Tagging}
\label{app:aspect_tagging}
To tag the focus topic within the peer review sentences, we implement a zero-shot aspect tagging procedure. We utilize \texttt{Llama-3.3-70B} to categorize each review sentence into predefined aspects. The model is configured with a temperature of 0.0 to favor deterministic outputs, and the maximum generation length is set to 512 tokens to accommodate the tag generation.

\section{Baseline Implementation}
\label{app:baseline_implementation}
We categorize the evaluated baselines into two groups: (1) specialized verification frameworks and (2) general-purpose LLM baselines. Tab.~\ref{tab:baseline_ckpt} lists the specific checkpoints and model versions used in our experiments for reproducibility. Prompts used for the LLM-based baselines are provided in Appendix~\ref{app:prompt_example}.

\noindent\textbf{Specialized Verification Frameworks.}
We evaluate four specialized hallucination verification frameworks: \texttt{HHEM-2.1-Open}~\cite{Baseline_hhem-2.1-open}, \texttt{True-NLI}~\cite{Baseline_TrueNLI_laurer2022less}, a \texttt{\textsc{seNtLI}}-style retriever-verifier pipeline~\citep{Baseline_SENTLI_schuster2022stretching}, and \texttt{RefChecker}~\cite{Baseline_hu2024refchecker}.

\texttt{HHEM-2.1-Open} produces a continuous faithfulness score between a review sentence and the corresponding source paper. \texttt{True-NLI} formulates hallucination detection as a natural language inference (NLI) task by estimating whether a review sentence is entailed by the paper content. For \texttt{True-NLI}, we compute entailment probabilities between each review sentence and all candidate paper chunks, and use the maximum entailment probability as the final verification score.

For the \texttt{\textsc{seNtLI}}-style pipeline\footnote{As the original \textsc{seNtLI} framework does not release a trained verifier, we implement this baseline using \texttt{RoBERTa-large-MNLI}.}, the system first retrieves relevant supporting or contradicting evidence from the source paper, followed by NLI-based verification conditioned on the retrieved evidence. Furthermore, we evaluate \texttt{RefChecker}~\cite{Baseline_hu2024refchecker}, a claim-level framework that operates on extracted claim triplets rather than full sentence representations. Following the original setup, we instantiate \texttt{RefChecker} using \texttt{Mistral-7B} for claim extraction alongside an AlignScore-based checker.

For all specialized verification frameworks, binary prediction thresholds are selected by maximizing F1 score on the training split and then fixed during test set evaluation. All inference procedures are conducted on a single NVIDIA H100 GPU.

\noindent\textbf{General-Purpose LLM Baselines.}
We evaluate both prompting-based and instruction-tuned LLM baselines for hallucination detection, hallucination type classification, and hallucination localization.

For prompting-based evaluation, we test \texttt{Qwen3-32B}, \texttt{Llama-3.3-70B}, \texttt{GPT-OSS-20B}, \texttt{GPT-OSS-120B}, \texttt{Mistral-Small-3.1-24B}, \texttt{RootSignals-Judge-Llama-70B}, and \texttt{GPT-5.2} in zero-shot settings.

We additionally implement a Retrieval-Augmented LLM-as-a-Judge (\texttt{RA-LLM}) framework based on \texttt{Qwen3-32B}. For each review sentence, the system retrieves relevant evidence chunks from the source paper and incorporates them into the prompt context for verification. Following prior hallucination evaluation work~\citep{HalluDetect_HaluEval_li2023}, we implement three prompting strategies: (1) \emph{Knowledge Retrieval (KR)}, which augments the prompt with demonstrations; (2) \emph{Chain-of-Thought (CoT)}, which encourages intermediate reasoning for consistency verification; and (3) \emph{Sample Contrast (Contrast)}, which prompts the model to first identify supporting and contradictory evidence before making the final verification decision.

For the instruction-tuned baselines, we perform supervised fine-tuning (SFT) using the \texttt{Unsloth} framework to optimize memory usage and training speed. We apply 4-bit NormalFloat (NF4) quantization \citep{dettmers2023qlora} on \texttt{Qwen2.5-3B-Instruct}, \texttt{Qwen2.5-7B-Instruct}, and \texttt{Qwen3-32B}.

We utilize QLoRA with a rank of $r=16$, an alpha parameter of $\alpha=32$, and a dropout rate of 0.05. To maximize representational capacity, the LoRA adapters are applied to all linear modules within the attention and MLP layers (\texttt{q\_proj, k\_proj, v\_proj, o\_proj, gate\_proj, up\_proj, down\_proj}), without adding bias terms. During training, we utilize the 8-bit AdamW optimizer with a peak learning rate of $2\times 10^{-4}$, decayed following a cosine schedule after a 5\% linear warmup. The models are trained in \texttt{bfloat16} precision with gradient checkpointing enabled. 

We maintain an effective batch size of 16 through gradient accumulation. To ensure the model focuses strictly on generation quality, we apply a completion-only loss masking strategy, computing the cross-entropy loss exclusively on the assistant's response tokens. The maximum sequence length is truncated at 4096 tokens. 

To ensure stable initial convergence, the early stopping mechanism is only activated after the completion of the first training epoch. Subsequently, we evaluate the models on a validation split (10\% of the training data) every $200$ steps and employ early stopping with a patience of $3$ evaluations based on the F1 score (after first epoch). During inference, all evaluations are conducted deterministically with the temperature set to 0.

All training and inference procedures are conducted on a single NVIDIA H100 GPU.

\begin{table*}[tb]
\small
\centering
\caption{Model identifiers and their corresponding checkpoints or versions used in this study.}
\label{tab:baseline_ckpt}
\begin{tabularx}{\textwidth}{lX}
\toprule
Baseline & Checkpoint \\
\midrule
HHEM-2.1-Open & \url{https://huggingface.co/vectara/hallucination_evaluation_model} \\
True-NLI & \url{https://huggingface.co/MoritzLaurer/DeBERTa-v3-large-mnli-fever-anli-ling-wanli} \\
SENTLI-style evidence retrieval & \url{https://huggingface.co/FacebookAI/roberta-large-mnli} \\
Refchecker (claim extractor) &
\url{https://huggingface.co/dongyru/Mistral-7B-Claim-Extractor} \\
\midrule
Qwen3-32B & \url{https://huggingface.co/Qwen/Qwen3-32B} \\
Llama-3.3-70B & \url{https://huggingface.co/meta-llama/Llama-3.3-70B-Instruct} \\
Mistral-Small-3.1-24B & \url{https://huggingface.co/mistralai/Mistral-Small-3.1-24B-Instruct-2503} \\
GPT-OSS-20B & \url{https://huggingface.co/openai/gpt-oss-20b} \\
GPT-OSS-120B & \url{https://huggingface.co/openai/gpt-oss-120b}\\
GPT-5.2 & \url{gpt-5.2-2025-12-11}\\
RootSignals-Judge-Llama-70B & \url{https://huggingface.co/root-signals/RootSignals-Judge-Llama-70B} \\
Qwen2.5-3B-Instruct & \url{https://huggingface.co/Qwen/Qwen2.5-3B-Instruct} \\
Qwen2.5-7B-Instruct & \url{https://huggingface.co/Qwen/Qwen2.5-7B-Instruct} \\
\bottomrule
\end{tabularx}
\end{table*}

\section{Evaluation Metrics}
\label{app:evaluation_metrics}
This appendix details the metrics used for the three tasks defined in Sec.~\ref{sec:eval}.

\subsection{Task 1: Hallucination Detection}
Accuracy, Precision, Recall, and F1 are computed using standard definitions. Given the class imbalance in hallucination labels, we additionally report the Matthews Correlation Coefficient (MCC), which accounts for all four entries of the confusion matrix and provides a more informative summary under skewed label distributions.

\subsection{Task 2: Hallucination Type Classification}
Evaluation is restricted to sentences annotated as hallucinated in the ground truth. We report two complementary metrics: (1) \textit{Macro-F1:} Computed by averaging F1 scores across all hallucination types, treating each class equally regardless of frequency. (2) \textit{Micro-F1:} Computed globally over all instances, reflecting overall classification accuracy weighted by class prevalence.

\subsection{Task 3: Hallucination Localization}
Gold and predicted spans are first aligned at the token level using the original tokenization of the review text.

\noindent\textbf{Token-level Evaluation.}
Gold and predicted spans are converted into BIO tag sequences, where both \texttt{B} and \texttt{I} tags are treated as positive labels and \texttt{O} as negative. Token-F1 is then computed over these binary labels.

\noindent\textbf{Span-level Evaluation.}
We report two complementary metrics: (1) \textit{Exact Match Span-F1:} Counts a predicted span as correct only if both its start and end boundaries exactly match a gold span. (2) \textit{Partial (Overlap) Span-F1:} Relaxes this criterion and counts a predicted span as correct if it overlaps with any gold span by at least one token. In cases of multiple predicted and gold spans within a sentence, matching is performed greedily to avoid double-counting.

\section{Additional Experiment Results}
\label{app:supp_experiment_result}
\subsection{Hallucination Detection (Task 1)}
\subsubsection{Ablation Study of RA-LLM}
\label{app:supp_experiment_result_ablation_RA-LLM}
In this section, we conduct an ablation study to isolate the impact of different evidence retrieval strategies on our retrieval-augmented LLM judge (\texttt{RA-LLM}). Specifically, we compare three configurations: (i) \textit{No-Retrieval}, where the judge directly consumes the full source paper content without any retrieval filtering; (ii) \textit{BM25}, where the judge is provided with paper content retrieved via BM25; and (iii) \textit{LCS}, where the judge receives the top-1 paper chunk containing the Longest Common Subsequence (LCS) with the target review sentence.
We report results under the same three prompting variants used in our main RA-LLM experiments (KR, CoT, and Contrast), in order to characterize how the impact of evidence differs across prompting designs.
Together, these ablations clarify the dominant failure modes in peer-review hallucination detection and quantify the upper bound achievable when perfect evidence is available.

Tab.~\ref{tab:ra_llm_ablation_neurips2024} reports the RA-LLM ablation on NeurIPS~2024 under three prompting variants. 

\noindent\textbf{Impact of Prompting Strategies.}
Among the RA-LLM variants, the choice of prompting strategy heavily influences the model's sensitivity to evidence retrieval. At the sentence level, \texttt{RA-LLM (KR)}, which augments prompts with few-shot demonstrations, consistently achieves the strongest performance (e.g., an MCC of 0.61 and Accuracy of 0.82 under the \texttt{BM25} setting), outperforming both explicit reasoning (\texttt{RA-LLM (CoT)}) and contrastive prompting (\texttt{RA-LLM (Contrast)}). This discrepancy can be attributed to the dependency of \texttt{CoT} and \texttt{Contrast} on the completeness of the source paper evidence. As observed in the table, transitioning from full-paper access (\texttt{No-Retrieval}) to filtered chunked evidence (\texttt{BM25} or \texttt{LCS}) often limits the context required for multi-step reasoning or contradiction analysis. Consequently, when the retrieved context lacks sufficient global information, the performance of \texttt{CoT} and \texttt{Contrast} strategies degrades (e.g., the review-level MCC of \texttt{Contrast} drops sharply from 0.26 under \texttt{No-Retrieval} to 0.15 under \texttt{BM25}). In this setting, the concise few-shot demonstrations in \texttt{RA-LLM (KR)} provide a more resilient and stable supervision signal, allowing the model to remain highly effective even with fragmented or localized evidence.

\begin{table}[ht] 
\centering
\setlength{\tabcolsep}{3pt}
\fontsize{6.5}{11}\selectfont
\caption{\textbf{Ablation of RA-LLM on HalluPeer (NeurIPS 2024).} Review-/sentence-level results are shown before/after the slash. For the third configuration, we use LCS as the chunking strategy at the review level, and Oracle at the sentence level.} 
\label{tab:ra_llm_ablation_neurips2024}
\begin{tabularx}{0.5\textwidth}{Xccccc}
\toprule
\textbf{Ablation Setting} & Acc. & Prec. & Rec. & F1 &  MCC \\
\cmidrule(lr){1-1}\cmidrule(lr){2-6}
\multicolumn{6}{l}{\textbf{RA-LLM (KR)}} \\
\midrule
\texttt{No-Retrieval} & 0.60 / 0.81 & 0.68 / 0.72 & 0.39 / 0.75 & 0.50 / 0.73 & 0.23 / 0.59 \\
\texttt{BM25} & \textbf{0.61} / \textbf{0.82} & \textbf{0.68} / \textbf{0.74} & \textbf{0.41} / 0.74 & \textbf{0.51} / \textbf{0.74} & \textbf{0.24} / \textbf{0.61} \\
\texttt{LCS} & 0.60 / 0.82 & 0.67 / 0.72 & 0.40 / \textbf{0.76} & 0.50 / 0.74 & 0.22 / 0.60 \\
\midrule
\midrule
\multicolumn{6}{l}{\textbf{RA-LLM (CoT)}} \\
\midrule
\texttt{No-Retrieval} & \textbf{0.62} / \textbf{0.78} & \textbf{0.71} / \textbf{0.66} & 0.42 / 0.73 & 0.53 / 0.70 & \textbf{0.27} / \textbf{0.53} \\
\texttt{BM25} & 0.61 / 0.77 & 0.63 / 0.63 & 0.52 / 0.78 & 0.57 / \textbf{0.70} & 0.22 / 0.53 \\
\texttt{LCS} & 0.60 / 0.76 & 0.61 / 0.61 & \textbf{0.56} / \textbf{0.80} & \textbf{0.58} / 0.70 & 0.20 / 0.52 \\
\midrule
\midrule
\multicolumn{6}{l}{\textbf{RA-LLM (Contrast)}} \\
\midrule
\texttt{No-Retrieval} & \textbf{0.63} / \textbf{0.75} & \textbf{0.64} / \textbf{0.61} & 0.61 / 0.74 & 0.62 / 0.67 & \textbf{0.26} / 0.48 \\
\texttt{BM25} & 0.57 / 0.73 & 0.55 / 0.57 & 0.77 / 0.84 & 0.64 / \textbf{0.68} & 0.15 / \textbf{0.49} \\
\texttt{LCS} & 0.58 / 0.72 & 0.56 / 0.56 & \textbf{0.80} / \textbf{0.85} & \textbf{0.66} / 0.68 & 0.18 / 0.48 \\
\bottomrule
\end{tabularx}
\end{table}

\subsubsection{In-Domain Results on ICLR 2024}
\label{app:supp_exp_in-domain_ICLR2024_task1}
Tab.~\ref{apptab:task1_detection} presents the performance of hallucination detection at both review and sentence levels on HalluPeer (ICLR 2024). The results are consistent with those on NeurIPS 2024, with the key findings in Sec.~\ref{sec:exp_result-task1} remaining unchanged.

Specialized verifiers again generalize poorly, with review-level $MCC \leq 0.04$ (near-random). RA-LLM (KR) remains the strongest prompting strategy at the sentence level (MCC 0.59), while prompting-based frontier LLMs continue to underperform at the review level relative to the sentence level. Domain-specific fine-tuning again dominates: even the compact Qwen2.5-3B surpasses all zero-shot baselines, and Qwen3-32B (Fine-tuned) achieves the best sentence-level performance (F1 0.94), with review-level F1 tied at 0.89.

\begin{table}[ht] 
    \centering
    \setlength{\tabcolsep}{3pt}
    \fontsize{6.05}{11}\selectfont
    \caption{\textbf{Task 1 results on HalluPeer (ICLR 2024).} Review-/sentence-level results are shown before/after the slash. Results are reported under the default evidence and prompting settings described in Sec.~\ref{subsec:baselines}.} 
    \label{apptab:task1_detection}
    \begin{tabularx}{0.5\textwidth}{Xccccc}
    \toprule
    \textbf{Specialized Verification} & Acc. & Prec. & Rec. & F1 &  MCC \\
    \cmidrule(lr){1-1}\cmidrule(lr){2-6}
    \texttt{HHEM-2.1-Open} & 0.51 / \textbf{0.63} & 0.51 / \textbf{0.48} & 0.61 / \textbf{0.65} & 0.56 / \textbf{0.55} & 0.03 / \textbf{0.25} \\
    \texttt{True-NLI} & 0.50 / 0.55 & 0.50 / 0.40 & 0.48 / 0.57 & 0.49 / 0.47 & 0.00 / 0.10 \\
    \texttt{\textsc{seNtLI}} & \textbf{0.52} / 0.60 & 0.51 / 0.45 & \textbf{0.63} / 0.58 & \textbf{0.57} / 0.51 & \textbf{0.04} / 0.18 \\
    \texttt{Refchecker} & \textbf{0.52} / 0.48 & \textbf{0.52} / 0.35 & 0.53 / 0.52 & 0.52 / 0.42 & \textbf{0.04} / -0.02 \\
    \midrule
    \midrule
    \textbf{LLM (Prompting)} & Acc. & Prec. & Rec. & F1 &  MCC \\
    \cmidrule(lr){1-1}\cmidrule(lr){2-6}
    \texttt{RA-LLM (KR)} & 0.59 / \textbf{0.81} & 0.59 / \textbf{0.72} & 0.59 / 0.76 & 0.59 / \textbf{0.74} & 0.18 / \textbf{0.59} \\
    \texttt{RA-LLM (CoT)} & 0.59 / 0.77 & 0.60 / 0.65 & 0.56 / 0.77 & 0.58 / 0.71 & 0.18 / 0.53 \\
    \texttt{RA-LLM (Contrast)} & 0.56 / 0.71 & 0.54 / 0.56 & 0.75 / \textbf{0.85} & 0.63 / 0.67 & 0.13 / 0.46 \\
    \texttt{Qwen3-32B} & 0.59 / 0.79 & 0.59 / 0.69 & 0.62 / 0.73 & 0.60 / 0.71 & 0.19 / 0.54 \\
    \texttt{Llama-3.3-70B} & 0.57 / 0.72 & \textbf{0.70} / 0.60 & 0.23 / 0.64 & 0.35 / 0.62 & 0.18 / 0.40 \\
    \texttt{Mistral-Small-3.1} & \textbf{0.61} / 0.74 & 0.65 / 0.61 & 0.48 / 0.72 & 0.55 / 0.66 & \textbf{0.22} / 0.45 \\
    \texttt{GPT-OSS-20B} & 0.57 / 0.78 & 0.55 / 0.67 & 0.74 / 0.74 & 0.63 / 0.71 & 0.15 / 0.53 \\
    \texttt{GPT-OSS-120B} & 0.55 / 0.79 & 0.53 / 0.67 & \textbf{0.87} / 0.79 & \textbf{0.66} / 0.73 & 0.13 / 0.56 \\
    \texttt{Judge-Llama-70B} & 0.56 / 0.71 & \textbf{0.70} / 0.59 & 0.23 / 0.65 & 0.34 / 0.62 & 0.17 / 0.39 \\
    \midrule
    \midrule
    \textbf{LLM (Fine-tuned)} & Acc. & Prec. & Rec. & F1 &  MCC \\
    \cmidrule(lr){1-1}\cmidrule(lr){2-6}
    \texttt{Qwen2.5-3B} & 0.85 / 0.92 & 0.84 / 0.85 & 0.86 / \textbf{0.92} & 0.85 / 0.89 & 0.70 / 0.82 \\
    \texttt{Qwen2.5-7B} & \textbf{0.90} / 0.94 & \textbf{0.93} / 0.91 & 0.85 / \textbf{0.92} & \textbf{0.89} / 0.92 & \textbf{0.79} / 0.87 \\
    \texttt{Qwen3-32B} & 0.89 / \textbf{0.96} & 0.90 / \textbf{0.96} & \textbf{0.88} / \textbf{0.92} & \textbf{0.89} / \textbf{0.94} & 0.78 / \textbf{0.90} \\
    \bottomrule
    \end{tabularx}
\end{table}


\subsection{Hallucination Type Classification (Task 2)}
\subsubsection{Per-label F1 Results}
\label{app:task2_per_label}
Tab.~\ref{tab:task2_per_label} presents both review-level and sentence-level per-label F1 results for hallucination category classification on HalluPeer (NeurIPS 2024).

\begin{table*}[ht] 
\centering
\setlength{\tabcolsep}{4pt}
\fontsize{7.8}{11}\selectfont
\caption{\textbf{Detailed Per-label F1 results for Task 2 on HalluPeer (NeurIPS 2024).} Review-/sentence-level results are shown before/after the slash. Per-label F1 abbreviations: A (\textit{Attribution Failure}), C (\textit{Context-based Meaning Error}), E (\textit{Entity}), F (\textit{False Concatenation}), H (\textit{Hyperbole}), N (\textit{Number}), O (\textit{Overgeneralization}), R (\textit{Reasoning Error}), T (\textit{Temporal}).}
\label{tab:task2_per_label}
\begin{tabularx}{\textwidth}{Xccccccccc}
\toprule
\textbf{LLM (Prompting)} & A & C & E & F & H & N & O & R & T \\
\cmidrule(lr){1-1}\cmidrule(lr){2-10}
\texttt{Qwen3-32B} & 0.06 / 0.15 & 0.03 / \textbf{0.29} & \textbf{0.30} / \textbf{0.41} & 0.12 / 0.26 & 0.00 / 0.05 & \textbf{0.40} / 0.55 & \textbf{0.36} / 0.23 & 0.10 / 0.32 & 0.00 / 0.25 \\
\texttt{Llama-3.3-70B} & 0.13 / 0.13 & 0.00 / 0.11 & 0.22 / 0.36 & 0.17 / 0.16 & 0.00 / 0.05 & 0.30 / 0.63 & 0.27 / 0.23 & 0.16 / 0.29 & 0.00 / 0.17 \\
\texttt{Mistral-Small-3.1} & 0.19 / 0.18 & 0.00 / 0.21 & 0.26 / 0.39 & \textbf{0.27} / \textbf{0.38} & \textbf{0.09} / 0.07 & 0.36 / \textbf{0.73} & 0.22 / \textbf{0.26} & \textbf{0.22} / \textbf{0.41} & \textbf{0.11} / 0.30 \\
\texttt{GPT-OSS-20B} & 0.04 / 0.17 & 0.03 / 0.12 & 0.19 / 0.34 & 0.07 / 0.21 & 0.07 / 0.07 & 0.30 / 0.61 & 0.27 / 0.21 & 0.16 / 0.32 & 0.00 / 0.15 \\
\texttt{GPT-OSS-120B} & \textbf{0.20} / \textbf{0.23} & 0.16 / 0.23 & 0.21 / 0.33 & 0.04 / 0.13 & 0.07 / 0.04 & 0.32 / 0.60 & 0.29 / 0.19 & 0.14 / 0.27 & 0.00 / 0.18 \\
\texttt{Judge-Llama-70B} & 0.14 / 0.13 & 0.00 / 0.12 & 0.21 / 0.35 & 0.19 / 0.16 & 0.00 / 0.04 & 0.33 / 0.60 & 0.31 / 0.24 & 0.10 / 0.28 & 0.00 / 0.14 \\
\texttt{GPT-5.2} & 0.00 / 0.19 & \textbf{0.17} / 0.19 & 0.20 / 0.34 & 0.11 / 0.12 & 0.07 / \textbf{0.15} & 0.33 / 0.66 & 0.19 / 0.23 & 0.18 / 0.38 & 0.00 / \textbf{0.31} \\
\midrule
\midrule
\textbf{LLM (Fine-tuned)} & A & C & E & F & H & N & O & R & T \\
\cmidrule(lr){1-1}\cmidrule(lr){2-10}
\texttt{Qwen2.5-3B} & 0.42 / 0.60 & \textbf{0.63} / 0.71 & 0.25 / 0.75 & 0.50 / 0.63 & \textbf{0.75} / 0.77 & 0.55 / 0.85 & 0.58 / 0.67 & 0.61 / 0.76 & 0.15 / 0.72 \\
\texttt{Qwen2.5-7B} & \textbf{0.56} / \textbf{0.77} & \textbf{0.63} / \textbf{0.84} & 0.40 / \textbf{0.84} & 0.49 / \textbf{0.83} & 0.63 / \textbf{0.89} & 0.51 / \textbf{0.91} & 0.58 / \textbf{0.89} & 0.61 / \textbf{0.88} & 0.35 / 0.85 \\
\texttt{Qwen3-32B} & 0.49 / 0.64 & 0.56 / 0.82 & \textbf{0.47} / 0.83 & \textbf{0.62} / 0.82 & 0.72 / 0.87 & \textbf{0.61} / \textbf{0.91} & \textbf{0.63} / 0.79 & \textbf{0.67} / 0.87 & \textbf{0.48} / \textbf{0.87} \\
\bottomrule
\end{tabularx}
\end{table*}

\subsubsection{In-Domain Results on ICLR 2024}
\label{app:supp_exp_in-domain_ICLR2024_task2}
Tabs.~\ref{apptab:task2_classification_overall} and
\ref{apptab:iclr-perlabel} present the overall and per-label F1 results on HalluPeer (ICLR 2024), respectively.
The results are consistent with those on NeurIPS 2024, with the key findings in Sec.~\ref{sec:exp_result-task2} remaining unchanged.

Zero-shot prompting remains weak, with the best review-level Macro-F1 reaching only 0.15, and larger models do not consistently outperform smaller ones (e.g., Mistral-Small-3.1 exceeds Llama-3.3-70B and GPT-OSS-120B). The sentence-vs.\ review-level gap persists, while fine-tuning yields substantial gains, particularly on semantically challenging categories such as Hyperbole, Temporal, and Context-based Meaning Error.



\begin{table}[h] 
    \centering
    \setlength{\tabcolsep}{10pt}
    \fontsize{8}{11}\selectfont
    \caption{\textbf{Task 2 overall results on HalluPeer (ICLR 2024).} Review-/sentence-level results are shown before/after the slash.}
    \label{apptab:task2_classification_overall}
    \begin{tabularx}{\columnwidth}{Xcc}
    \toprule
    \textbf{LLM (Prompting)} & Macro-F1 & Micro-F1 \\
    \cmidrule(lr){1-1}\cmidrule(lr){2-3}
    \texttt{Qwen3-32B} & 0.09 / 0.29 & 0.15 / 0.31 \\
    \texttt{Llama-3.3-70B} & 0.12 / 0.24 & 0.16 / 0.27 \\
    \texttt{Mistral-Small-3.1} & \textbf{0.15} / \textbf{0.33} & \textbf{0.22} / \textbf{0.36} \\
    \texttt{GPT-OSS-20B} & 0.13 / 0.26 & 0.17 / 0.29 \\
    \texttt{GPT-OSS-120B} & 0.12 / 0.25 & 0.16 / 0.26 \\
    \texttt{Judge-Llama-70B} & 0.13 / 0.23 & 0.17 / 0.26 \\
    \midrule
    \midrule
    \textbf{LLM (Fine-tuned)} & Macro-F1 & Micro-F1 \\
    \cmidrule(lr){1-1}\cmidrule(lr){2-3}
    \texttt{Qwen2.5-3B} & 0.53 / 0.65 & 0.60 / 0.67 \\
    \texttt{Qwen2.5-7B} & \textbf{0.60} / 0.82 & 0.60 / 0.82 \\
    \texttt{Qwen3-32B} & 0.59 / \textbf{0.85} & \textbf{0.63} / \textbf{0.86} \\
    \bottomrule
    \end{tabularx}
\end{table}

\begin{table*}[ht] 
\centering
\setlength{\tabcolsep}{4pt}
\fontsize{7.8}{11}\selectfont
\caption{\textbf{Detailed Per-label F1 results for Task 2 on HalluPeer (ICLR 2024).} Review-/sentence-level results are shown before/after the slash. Per-label F1 abbreviations: A (\textit{Attribution Failure}), C (\textit{Context-based Meaning Error}), E (\textit{Entity}), F (\textit{False Concatenation}), H (\textit{Hyperbole}), N (\textit{Number}), O (\textit{Overgeneralization}), R (\textit{Reasoning Error}), T (\textit{Temporal}).}
\label{apptab:iclr-perlabel}
\begin{tabularx}{\textwidth}{Xccccccccc}
\toprule
\textbf{LLM (Prompting)} & A & C & E & F & H & N & O & R & T \\
\cmidrule(lr){1-1}\cmidrule(lr){2-10}
\texttt{Qwen3-32B} & 0.06 / 0.22 & 0.05 / \textbf{0.32} & \textbf{0.24} / \textbf{0.41} & 0.16 / 0.28 & 0.00 / 0.04 & 0.00 / 0.53 & 0.11 / 0.21 & 0.22 / 0.34 & 0.00 / 0.24 \\
\texttt{Llama-3.3-70B} & 0.11 / 0.17 & 0.03 / 0.16 & 0.18 / 0.35 & 0.05 / 0.19 & \textbf{0.08} / 0.03 & 0.18 / 0.63 & 0.17 / 0.21 & 0.29 / 0.30 & 0.00 / 0.14 \\
\texttt{Mistral-Small-3.1} & 0.15 / 0.23 & 0.05 / 0.27 & 0.20 / 0.41 & \textbf{0.28} / \textbf{0.39} & 0.00 / \textbf{0.06} & \textbf{0.26} / \textbf{0.71} & 0.14 / \textbf{0.24} & \textbf{0.29} / \textbf{0.42} & 0.00 / \textbf{0.28} \\
\texttt{GPT-OSS-20B} & 0.21 / \textbf{0.24} & \textbf{0.08} / 0.19 & 0.17 / 0.31 & 0.09 / 0.18 & 0.00 / 0.05 & 0.18 / 0.60 & 0.16 / 0.23 & 0.27 / 0.36 & 0.00 / 0.15 \\
\texttt{GPT-OSS-120B} & \textbf{0.32} / 0.23 & 0.05 / 0.30 & 0.16 / 0.32 & 0.00 / 0.13 & 0.00 / 0.04 & 0.20 / 0.59 & 0.15 / 0.17 & 0.24 / 0.30 & 0.00 / 0.22 \\
\texttt{Judge-Llama-70B} & 0.15 / 0.16 & 0.05 / 0.17 & 0.20 / 0.34 & 0.05 / 0.19 & \textbf{0.08} / 0.03 & 0.16 / 0.61 & \textbf{0.22} / 0.21 & 0.25 / 0.26 & 0.00 / 0.14 \\
\midrule
\midrule
\textbf{LLM (Fine-tuned)} & A & C & E & F & H & N & O & R & T \\
\cmidrule(lr){1-1}\cmidrule(lr){2-10}
\texttt{Qwen2.5-3B} & 0.44 / 0.50 & 0.62 / 0.65 & \textbf{0.50} / 0.66 & 0.54 / 0.64 & 0.54 / 0.67 & 0.42 / 0.81 & 0.62 / 0.51 & 0.70 / 0.73 & 0.34 / 0.68 \\
\texttt{Qwen2.5-7B} & \textbf{0.44} / 0.72 & 0.63 / 0.79 & 0.48 / 0.83 & 0.57 / 0.79 & \textbf{0.71} / 0.87 & 0.50 / 0.92 & \textbf{0.79} / 0.81 & 0.61 / 0.86 & \textbf{0.65} / 0.84 \\
\texttt{Qwen3-32B} & 0.42 / \textbf{0.75} & \textbf{0.68} / \textbf{0.82} & 0.42 / \textbf{0.87} & \textbf{0.58} / \textbf{0.82} & 0.68 / \textbf{0.88} & \textbf{0.57} / \textbf{0.93} & 0.76 / \textbf{0.84} & \textbf{0.73} / \textbf{0.89} & 0.53 / \textbf{0.87} \\
\bottomrule
\end{tabularx}
\end{table*}

\subsection{Hallucination Localization (Task 3)}
\subsubsection{In-Domain Results on ICLR 2024}
\label{app:supp_exp_in-domain_ICLR2024_task3}

Tab.~\ref{apptab:task3_localization} presents the results for hallucination span localization on HalluPeer (ICLR 2024). The results are consistent with those on NeurIPS 2024, with the key findings in Sec.~\ref{sec:exp_result-task3} remaining unchanged. 

Zero-shot span localization remains limited, with the consistent gap between Overlap and Exact Span-F1 also observed. Fine-tuning substantially improves grounding, with Qwen3-32B (Fine-tuned) achieving 0.93 Token-F1 and 0.90 Exact Span-F1.

\begin{table}[ht] 
    \centering
    \setlength{\tabcolsep}{4pt}
    \fontsize{8}{11}\selectfont
    \caption{\textbf{Task 3 results on HalluPeer (ICLR 2024).} Review-level results are reported under default evidence retrieval and prompting settings described in Sec.~\ref{subsec:baselines}.}
    \label{apptab:task3_localization}
    \begin{tabularx}{0.5\textwidth}{Xccc}
    \toprule
    \textbf{LLM (Prompting)} & Token-F1 & Exact Span-F1 & Overlap Span-F1 \\
    \cmidrule(lr){1-1}\cmidrule(lr){2-4}
    \texttt{Qwen3-32B} & 0.41 & 0.25 & 0.44 \\
    \texttt{Llama-3.3-70B} & 0.40 & 0.25 & 0.40 \\
    \texttt{Mistral-Small-3.1} & 0.34 & 0.20 & 0.34 \\
    \texttt{GPT-OSS-20B} & 0.48 & 0.33 & 0.49 \\
    \texttt{GPT-OSS-120B} & \textbf{0.54} & \textbf{0.43} & \textbf{0.54} \\
    \texttt{Judge-Llama-70B} & 0.41 & 0.28 & 0.40 \\
    \midrule
    \midrule
    \textbf{LLM (Fine-tuned)} & Token-F1 & Exact Span-F1 & Overlap Span-F1 \\
    \cmidrule(lr){1-1}\cmidrule(lr){2-4}
    \texttt{Qwen2.5-3B} & 0.84 & 0.77 & 0.82 \\
    \texttt{Qwen2.5-7B} & 0.87 & 0.83 & 0.86 \\
    \texttt{Qwen3-32B} & \textbf{0.93} & \textbf{0.90} & \textbf{0.92} \\
    \bottomrule
    \end{tabularx}
\end{table}

\subsection{Cross-Venue Transferability}
\label{app:cross_venue}
This section presents the full quantitative results (Tab.~\ref{tab:cross_venue_task1}--\ref{tab:cross_venue_task3}) corresponding to the analysis in Sec.~\ref{sec:cross-venue}.

\begin{table}[ht]
\centering
\setlength{\tabcolsep}{4pt}
\fontsize{6.55}{11}\selectfont
\caption{\textbf{Task 1 (Detection) cross-venue transfer.}}
\label{tab:cross_venue_task1}
\begin{tabularx}{\linewidth}{l *{5}{>{\centering\arraybackslash}X}}
\toprule
Model & Acc. & Prec. & Rec. & F1 & MCC \\
\midrule
\rowcolor{gray!15} \multicolumn{6}{c}{\textbf{\textit{Train:} NeurIPS 2024 $\rightarrow$ \textit{Test:} ICLR 2024}} \\
Qwen2.5-3B & 0.84 / 0.85 & 0.83 / 0.73 & 0.85 / \textbf{0.95} & 0.84 / 0.82 & 0.68 / 0.72 \\
Qwen2.5-7B & 0.88 / 0.91 & 0.88 / 0.83 & \textbf{0.89} / 0.93 & 0.88 / 0.88 & 0.76 / 0.81 \\
Qwen3-32B & \textbf{0.91} / \textbf{0.94} & \textbf{0.97} / \textbf{0.94} & 0.85 / 0.89 & \textbf{0.90} / \textbf{0.91} & \textbf{0.83} / \textbf{0.87} \\
\midrule
\rowcolor{gray!15} \multicolumn{6}{c}{\textbf{\textit{Train:} ICLR 2024 $\rightarrow$ \textit{Test:} NeurIPS 2024}} \\
Qwen2.5-3B & 0.82 / 0.92 & 0.81 / 0.85 & 0.83 / 0.91 & 0.82 / 0.88 & 0.63 / 0.82 \\
Qwen2.5-7B & \textbf{0.88} / 0.94 & \textbf{0.89} / 0.89 & \textbf{0.87} / \textbf{0.92} & \textbf{0.88} / 0.91 & \textbf{0.76} / 0.86 \\
Qwen3-32B & 0.86 / \textbf{0.95} & 0.87 / \textbf{0.94} & 0.85 / 0.91 & 0.86 / \textbf{0.93} & 0.72 / \textbf{0.89} \\
\bottomrule
\end{tabularx}
\end{table}

\begin{table}[ht]
\centering
\setlength{\tabcolsep}{5pt}
\fontsize{8}{11}\selectfont
\caption{\textbf{Task 2 (Type Classification) cross-venue transfer.}}
\label{tab:cross_venue_task2_overall}
\begin{tabularx}{\linewidth}{l *{2}{>{\centering\arraybackslash}X}}
\toprule
Model & Macro-F1 & Micro-F1 \\
\midrule
\rowcolor{gray!15} \multicolumn{3}{c}{\textbf{\textit{Train:} NeurIPS 2024 $\rightarrow$ \textit{Test:} ICLR 2024}} \\
Qwen2.5-3B & 0.52 / 0.73 & 0.56 / 0.73 \\
Qwen2.5-7B & \textbf{0.63} / \textbf{0.86} & 0.64 / \textbf{0.86} \\
Qwen3-32B & 0.62 / 0.82 & \textbf{0.66} / 0.83 \\
\midrule
\rowcolor{gray!15} \multicolumn{3}{c}{\textbf{\textit{Train:} ICLR 2024 $\rightarrow$ \textit{Test:} NeurIPS 2024}} \\
Qwen2.5-3B & 0.51 / 0.65 & 0.54 / 0.68 \\
Qwen2.5-7B & 0.53 / 0.81 & 0.57 / 0.81 \\
Qwen3-32B & \textbf{0.55} / \textbf{0.85} & \textbf{0.59} / \textbf{0.86} \\
\bottomrule
\end{tabularx}
\end{table}

\begin{table}[ht]
\centering
\setlength{\tabcolsep}{5pt}
\fontsize{8}{11}\selectfont
\caption{\textbf{Task 3 (Localization) cross-venue transfer.}}
\label{tab:cross_venue_task3}
\begin{tabular*}{\linewidth}{l@{\extracolsep{\fill}}ccc}
\toprule
Model & Token-F1 & Exact Span-F1 & Overlap Span-F1 \\
\midrule
\rowcolor{gray!15} \multicolumn{4}{c}{\textbf{\textit{Train:} NeurIPS 2024 $\rightarrow$ \textit{Test:} ICLR 2024}} \\
Qwen2.5-3B & 0.89 & 0.85 & 0.89 \\
Qwen2.5-7B & 0.88 & 0.85 & 0.88 \\
Qwen3-32B & \textbf{0.91} & \textbf{0.88} & \textbf{0.91} \\
\midrule
\rowcolor{gray!15} \multicolumn{4}{c}{\textbf{\textit{Train:} ICLR 2024 $\rightarrow$ \textit{Test:} NeurIPS 2024}} \\
Qwen2.5-3B & 0.78 & 0.67 & 0.73 \\
Qwen2.5-7B & 0.87 & 0.83 & 0.86 \\
Qwen3-32B & \textbf{0.92} & \textbf{0.87} & \textbf{0.91} \\
\bottomrule
\end{tabular*}
\end{table}


\subsection{Cross-Generation Ablation}
\label{app:cross_generation}

\noindent\textbf{Task 1 (Detection).} As shown in Tab.~\ref{tab:R1W2_cross_t1_merged}, the performance shift ($\Delta$) when testing on cross-generator data is extremely marginal. For the Mistral injector, the F1 score decreases by an average of only 0.01 to 0.02. Remarkably, for the Llama-3.3-70B injector, the $\Delta$ for both F1 and MCC is entirely non-negative across all splits, indicating that performance is preserved and even slightly improved on the unseen generator.

\noindent\textbf{Task 2 (Type Classification).} Tab.~\ref{tab:R1W2_cross_t2_merged} illustrates that classification performance remains highly stable, with shifts generally contained within 0.05. Notably, the review-level Micro-F1 improves under the Llama injector on both splits (up to +0.066), while the sentence-level performance exhibits only minor degradations.

\noindent\textbf{Task 3 (Localization).} As detailed in Tab.~\ref{tab:R1W2_cross_t3_merged}, the boundary grounding capabilities of our detector transfer robustly to unseen generators. The Token-F1 scores remain within 0.02 of the in-domain baseline for Mistral and show slight improvements on the ICLR split for Llama.

These findings indicate that our fine-tuned Qwen3-32B detector does not merely exploit spurious lexical artifacts or stylistic tics. Instead, it captures the semantic inconsistencies that define peer-review hallucinations, rather than exploiting superficial lexical artifacts.

\begin{table*}[t]
    \centering \small
    \setlength{\tabcolsep}{4pt}
    \renewcommand{\arraystretch}{1.2}
    \resizebox{\linewidth}{!}{
    \begin{tabular}{l ccc ccc ccc ccc ccc}
    \toprule
    \multirow{2}{*}{\textbf{Venue}}
    & \multicolumn{3}{c}{Accuracy} & \multicolumn{3}{c}{Precision}
    & \multicolumn{3}{c}{Recall} & \multicolumn{3}{c}{F1} & \multicolumn{3}{c}{MCC} \\
    \cmidrule(lr){2-4}\cmidrule(lr){5-7}\cmidrule(lr){8-10}\cmidrule(lr){11-13}\cmidrule(lr){14-16}
    & In & Cross & $\Delta\downarrow$ & In & Cross & $\Delta\downarrow$
    & In & Cross & $\Delta\downarrow$ & In & Cross & $\Delta\downarrow$ & In & Cross & $\Delta\downarrow$ \\
    \midrule
    \rowcolor{gray!15} \multicolumn{16}{c}{\textbf{Hallucination Injector: Mistral-Small-3.1}} \\
    NeurIPS & 90.4/94.2 & 88.7/93.5 & $-1.6$/$-0.7$ & 95.6/93.8 & 95.5/93.1 & $-0.1$/$-0.7$ & 84.6/88.6 & 81.2/87.4 & $-3.3$/$-1.3$ & 89.8/91.1 & 87.8/90.1 & $-2.0$/$-1.0$ & 81.3/86.8 & 78.3/85.4 & $-2.9$/$-1.5$ \\
    ICLR    & 89.2/95.5 & 88.4/95.2 & $-0.8$/$-0.3$ & 89.9/95.5 & 92.2/95.8 & $+2.3$/$+0.3$ & 88.3/91.8 & 83.9/90.5 & $-4.4$/$-1.3$ & 89.1/93.6 & 87.9/93.1 & $-1.2$/$-0.5$ & 78.4/90.2 & 77.1/89.5 & $-1.3$/$-0.7$ \\
    \midrule
    \rowcolor{gray!15} \multicolumn{16}{c}{\textbf{Hallucination Injector: Llama-3.3-70B}} \\
    NeurIPS & 90.4/94.2 & 92.1/94.9 & $+1.7$/$+0.8$ & 95.6/93.8 & 96.1/94.0 & $+0.5$/$+0.2$ & 84.6/88.6 & 87.7/91.0 & $+3.1$/$+2.4$ & 89.8/91.1 & 91.7/92.5 & $+1.9$/$+1.4$ & 81.3/86.8 & 84.5/88.7 & $+3.2$/$+1.8$ \\
    ICLR    & 89.2/95.5 & 89.3/95.9 & $+0.1$/$+0.3$ & 89.9/95.5 & 90.0/95.1 & $+0.1$/$-0.4$ & 88.3/91.8 & 88.4/93.2 & $+0.1$/$+1.4$ & 89.1/93.6 & 89.2/94.2 & $+0.1$/$+0.6$ & 78.4/90.2 & 78.7/91.0 & $+0.2$/$+0.8$ \\
    \bottomrule
    \end{tabular}}
    \caption{\textbf{Task~1 (Detection) cross-generator ablation --- Hallucination Injectors: Mistral-Small-3.1 and Llama-3.3-70B.} The Qwen3-32B detector is fine-tuned only on Qwen-injected data and evaluated on Qwen-injected (\textit{In}-domain) vs.\ Mistral- or Llama-injected (\textit{Cross}) test sets. Venue refers to the HalluPeer NeurIPS 2024 and ICLR 2024 splits. Values are presented as review-level\,/\,sentence-level (in \%). $\Delta=\text{Cross}-\text{In}$; $\downarrow$ indicates smaller $|\Delta|$ is better.}
    \label{tab:R1W2_cross_t1_merged}
\end{table*}

\begin{table}[t]
    \centering 
    \setlength{\tabcolsep}{2pt}
    \fontsize{6.8}{11}\selectfont
    \renewcommand{\arraystretch}{1.2}
    \begin{tabularx}{0.5\textwidth}{l ccc ccc}
    \toprule
    \multirow{2}{*}{\textbf{Venue}}
    & \multicolumn{3}{c}{Macro-F1} & \multicolumn{3}{c}{Micro-F1} \\
    \cmidrule(lr){2-4}\cmidrule(lr){5-7}
    & In & Cross & $\Delta\downarrow$ & In & Cross & $\Delta\downarrow$ \\
    \midrule
    \rowcolor{gray!15} \multicolumn{7}{c}{\textbf{Hallucination Injector: Mistral-Small-3.1}} \\
    NeurIPS & 58.5/82.4 & 61.4/79.2 & $+2.9$/$-3.2$ & 59.3/83.8 & 64.5/81.1 & $+5.2$/$-2.6$ \\
    ICLR    & 59.5/85.3 & 55.7/84.8 & $-3.8$/$-0.5$ & 62.8/85.6 & 60.4/85.0 & $-2.4$/$-0.6$ \\
    \midrule
    \rowcolor{gray!15} \multicolumn{7}{c}{\textbf{Hallucination Injector: Llama-3.3-70B}} \\
    NeurIPS & 58.5/82.4 & 62.4/76.5 & $+3.9$/$-5.9$ & 59.3/83.8 & 65.8/78.9 & $+6.6$/$-4.9$ \\
    ICLR    & 59.5/85.3 & 57.6/83.7 & $-1.8$/$-1.6$ & 62.8/85.6 & 64.4/84.5 & $+1.6$/$-1.1$ \\
    \bottomrule
    \end{tabularx}
    \caption{\textbf{Task~2 (Type Classification) cross-generator ablation --- Hallucination Injectors: Mistral-Small-3.1 and Llama-3.3-70B.} The Qwen3-32B detector is fine-tuned only on Qwen-injected data and evaluated on Qwen-injected (\textit{In}-domain) vs.\ Mistral- or Llama-injected (\textit{Cross}) test sets. Venue refers to the HalluPeer NeurIPS 2024 and ICLR 2024 splits. Values are presented as review-level\,/\,sentence-level (in \%). $\Delta=\text{Cross}-\text{In}$; $\downarrow$ indicates smaller $|\Delta|$ is better.}
    \label{tab:R1W2_cross_t2_merged}
\end{table}

\begin{table}[t]
    \centering \small
    \setlength{\tabcolsep}{2.2pt}
    \fontsize{8}{11}\selectfont
    \renewcommand{\arraystretch}{1.2}
    \begin{tabularx}{0.5\textwidth}{l ccc ccc ccc}
    \toprule
    \multirow{2}{*}{\textbf{Venue}}
    & \multicolumn{3}{c}{Token-F1} & \multicolumn{3}{c}{Exact Span-F1} & \multicolumn{3}{c}{Overlap Span-F1} \\
    \cmidrule(lr){2-4}\cmidrule(lr){5-7}\cmidrule(lr){8-10}
    & In & Cross & $\Delta\downarrow$ & In & Cross & $\Delta\downarrow$ & In & Cross & $\Delta\downarrow$ \\
    \midrule
    \rowcolor{gray!15} \multicolumn{10}{c}{\textbf{Hallucination Injector: Mistral-Small-3.1}} \\
    NeurIPS & 90.6 & 88.8 & $-1.8$ & 85.6 & 81.0 & $-4.5$ & 89.8 & 86.6 & $-3.2$ \\
    ICLR    & 92.8 & 92.3 & $-0.4$ & 89.6 & 87.6 & $-2.0$ & 92.3 & 90.4 & $-1.9$ \\
    \midrule
    \rowcolor{gray!15} \multicolumn{10}{c}{\textbf{Hallucination Injector: Llama-3.3-70B}} \\
    NeurIPS & 90.6 & 89.2 & $-1.4$ & 85.6 & 84.6 & $-1.0$ & 89.8 & 86.8 & $-3.0$ \\
    ICLR    & 92.8 & 94.8 & $+2.0$ & 89.6 & 92.2 & $+2.7$ & 92.3 & 93.5 & $+1.2$ \\
    \bottomrule
    \end{tabularx}
    \caption{\textbf{Task~3 (Localization) cross-generator ablation --- Hallucination Injectors: Mistral-Small-3.1 and Llama-3.3-70B.} The Qwen3-32B detector is fine-tuned only on Qwen-injected data and evaluated on Qwen-injected (\textit{In}-domain) vs.\ Mistral- or Llama-injected (\textit{Cross}) test sets. Venue refers to the HalluPeer NeurIPS 2024 and ICLR 2024 splits. Values are presented at the review-level (in \%), as Task~3 has no sentence-level split. $\Delta=\text{Cross}-\text{In}$; $\downarrow$ indicates smaller $|\Delta|$ is better.}
\label{tab:R1W2_cross_t3_merged}
\end{table}
\section{Details of the Taxonomy Generation and Refinement Pipeline}
\label{app:ensemble_details}
To reduce model-specific biases introduced during recursive taxonomy decomposition, we implement a three-stage refinement pipeline consisting of multi-model ensemble generation, automated overlap identification, and human expert validation. As illustrated in Fig.~\ref{fig:ensemble_pipeline}, the pipeline progressively refines the initially generated taxonomy into a globally consistent and human-validated hierarchy.

\subsection{Stage 1: Multi-Model Taxonomy Ensemble}
Starting from the predefined coarse-grained anchors (\textit{e.g.}, \emph{Number}, \emph{Entity}), we independently generate taxonomy trees using three LLM proposers: \texttt{Qwen3-32B}, \texttt{Llama-3.3-70B}, and \texttt{Mistral-Small-3.1-24B}. The taxonomy generated by \texttt{Qwen3-32B} is treated as the primary structure.

To determine whether a concept generated in the primary tree is supported by other model-generated trees, we concatenate each node's concept name and operational description into a single textual representation and encode it using the \texttt{BAAI/bge-large-en-v1.5} embedding model. For each node in the primary tree, we compute cosine similarity scores against all nodes in the other trees. A node is retained only if at least one cross-tree node achieves a cosine similarity score greater than 0.8. This cross-model agreement criterion helps reduce model-specific artifacts and improves the robustness of the induced taxonomy structure.

\subsection{Stage 2: Global Overlap Identification}
Although local decomposition constraints encourage non-overlapping sibling categories, semantically redundant concepts may still emerge across different branches of the taxonomy. To improve global mutual exclusivity, we perform a taxonomy-wide overlap identification stage.

We first construct a candidate comparison pool by enumerating all possible node pairs across the taxonomy, excluding direct parent--child node pairs. Exhaustive human inspection over this candidate space would result in an impractically large number of comparisons (approximately 86,800 node pairs). Therefore, we first apply a multi-LLM consensus filtering procedure in which multiple evaluator models independently assess whether two taxonomy nodes exhibit substantial semantic overlap. Only node pairs identified by consensus are forwarded for manual review. This process reduces the candidate pool to 434 potentially overlapping node pairs while mitigating blind spots introduced by any individual evaluator model.

\subsection{Stage 3: Human Expert Validation}
In the final stage, human annotators manually review the filtered set of candidate overlap pairs. Annotators inspect the semantic concepts, operational descriptions, and taxonomy paths associated with each node pair, and determine whether the nodes should be merged, preserved as distinct concepts, or revised for clearer separation.

The core judgment criteria used during human validation are described in Fig.~\ref{fig:criteria_hallucination_taxonomy_refinement}. This human-in-the-loop verification step helps eliminate residual redundancy and improve the conceptual consistency of the final taxonomy.

\section{Evaluation on Authentic Reviews}
\label{app:real_review_quantitative}
To quantitatively evaluate whether detectors trained on synthetic hallucinations transfer to naturally occurring reviewer errors, we construct a manually annotated set of authentic NeurIPS 2024 reviews. The evaluation is designed to avoid model-dependent annotation and to measure both true-positive and false-positive behavior.

\subsection{Setup}
Two expert annotators manually examined 1,161 real NeurIPS 2024 reviews against their corresponding submissions and identified 20 reviewer hallucinations, corresponding to a natural prevalence of approximately 1.7\%. Unlike the case study in Sec.~\ref{sec:sim_to_real}, the annotation was performed independently of any detector output: annotators read each review directly and determined which instances constitute genuine reviewer hallucinations.

The fine-tuned detectors were trained exclusively on synthetic HalluPeer ICLR 2024 data and had no access to the manually annotated authentic reviews. We evaluate the resulting detectors directly on the 1,161 authentic reviews, making this an out-of-distribution evaluation from synthetic hallucinations to naturally occurring reviewer errors.

\subsection{Results and Analysis}
\noindent\textbf{True and False Positive Rates.}
Table~\ref{tab:real_review_quantitative} reports the true-positive rate (TPR) and false-positive rate (FPR) on the manually annotated authentic reviews. Given the low natural prevalence of hallucinations, recall is the primary measure of detection capability, while FPR characterizes the amount of reviewer-level screening required.

Our fine-tuned Qwen3-32B detector recovers all 20 authentic hallucinations, achieving TPR $=100.0\%$ at FPR $=22.1\%$. This substantially improves recall over the zero-shot Qwen3-32B baseline (TPR $=95.0\%$), while Qwen2.5-7B reaches TPR $=70.0\%$. Among the compared frontier-model prompting baselines, Claude Opus 4.7 achieves TPR $=85.0\%$ at FPR $=0.3\%$.

\noindent\textbf{Effect of Fine-tuning.}
To isolate the effect of training on HalluPeer, we compare each fine-tuned detector against its corresponding zero-shot model. Fine-tuning Qwen3-32B improves TPR from 95.0\% to 100.0\% while reducing FPR from 29.5\% to 22.1\%. Similarly, Qwen2.5-7B improves TPR from 70.0\% to 80.0\% and reduces FPR from 34.0\% to 23.1\%. The simultaneous improvement in TPR and reduction in FPR indicates that the gain is not explained by a simple shift in the decision threshold.

\noindent\textbf{Practical Screening.}
At the reported operating point, the fine-tuned Qwen3-32B detector reduces the 1,161 authentic reviews to 272 candidates while recovering all 20 annotated hallucinations. Thus, the detector can serve as a high-recall screening stage, substantially narrowing the candidate pool while retaining the authentic hallucinations identified by expert annotators. Final adjudication can then be performed by a human or a stronger frontier model.

\begin{table}[ht]
\centering
\setlength{\tabcolsep}{5pt}
\fontsize{8}{11}\selectfont
\caption{\textbf{Detection performance on manually annotated real NeurIPS 2024 reviews.}
The evaluation set contains 1,161 reviews with 20 positive and 1,141 negative instances (1.7\% prevalence). Fine-tuned models are trained on synthetic HalluPeer ICLR 2024 data only. Values are reported in \%.}
\label{tab:real_review_quantitative}
\begin{tabularx}{\linewidth}{l *{4}{>{\centering\arraybackslash}X}}
\toprule
\textbf{Model} & \textbf{TPR} & \textbf{FPR} & \textbf{TP/FN} & \textbf{FP/TN} \\
\midrule
\rowcolor{gray!15} \multicolumn{5}{c}{\textit{Frontier LLM (prompting)}} \\
Claude Opus 4.7 & 85.0 & 0.3 & 17/3 & 3/1138 \\
\midrule
\rowcolor{gray!15} \multicolumn{5}{c}{\textit{Open-source, zero-shot}} \\
GPT-OSS-120B & 100.0 & 33.0 & 20/0 & 376/765 \\
Qwen3-32B & 95.0 & 29.5 & 19/1 & 337/804 \\
Qwen2.5-7B & 70.0 & 34.0 & 14/6 & 388/753 \\
\midrule
\rowcolor{gray!15} \multicolumn{5}{c}{\textit{Open-source, fine-tuned on HalluPeer}} \\
Qwen3-32B & \textbf{100.0} & 22.1 & 20/0 & 252/889 \\
Qwen2.5-7B & 80.0 & \textbf{23.1} & 16/4 & 264/877 \\
\bottomrule
\end{tabularx}
\end{table}
\section{Annotation Protocol for Authentic Review Analysis}
\label{app:real_review_annotation}
We selected 200 consecutive flagged instances from the 13,803 human-written NeurIPS 2024 reviews identified by our fine-tuned detector for manual inspection. Two expert annotators each reviewed a disjoint subset, checking whether each flagged instance constituted a genuine hallucination and whether the predicted hallucination type matched the manually assigned type. Candidate hallucinations and uncertain cases were subsequently reviewed jointly with a senior expert, with final labels determined through discussion and consensus. This process yielded 11 validated cases, which are presented in Tab.~\ref{tab:hallu-taxonomy-examples}--\ref{tab:hallu-taxonomy-examples-3}.

\section{Algorithm}
\label{app:algorithm}

\subsection{Algorithm -- Hallucination Taxonomy Construction}
Algorithm~\ref{alg:taxonomy_decomposition} summarizes the recursive procedure used to build the hallucination taxonomy. Starting from the predefined first-level concepts $\mathcal{V}^{(1)}$, the algorithm performs a depth-first traversal that progressively refines each coarse concept into finer subtypes. For a node at depth $d$, if the maximum depth $D$ has not been reached, the LLM proposer $\textsc{Expand}_{\mathcal{M}}$ is queried to generate a set of child concepts $\mathcal{C}_v$ conditioned on the node and its description $\delta_v$. Each generated child is attached to its parent, assigned an operational description via $\textsc{Describe}_{\mathcal{M}}$, added to the global node set $\mathcal{V}$, and then recursively decomposed at depth $d+1$. The recursion terminates under two conditions: when the maximum depth $D$ is reached, or when $\textsc{Expand}_{\mathcal{M}}$ returns no children ($\mathcal{C}_v = \emptyset$), indicating that the concept is atomic. The procedure yields the full taxonomy tree $\mathcal{V}$, whose leaf nodes serve as the fine-grained hallucination types used for injection.

\begin{algorithm}[t]
\footnotesize
\SetInd{0.4em}{0.7em}
\caption{LLM-Guided Recursive Taxonomy Decomposition (DFS)}
\label{alg:taxonomy_decomposition}

\KwIn{Root node $v^{(0)}$, Predefined first-level nodes $\mathcal{V}^{(1)}$, LLM $\mathcal{M}$, Max depth $D$}
\KwOut{Full taxonomy tree $\mathcal{V}$}

$\mathcal{V} \leftarrow \{v^{(0)}\} \cup \mathcal{V}^{(1)}$\;
\ForEach{$v \in \mathcal{V}^{(1)}$}{
    \tcp{Start DFS from each first-level node}
    $\text{Decompose}(v, 1)$\;
}
\Return $\mathcal{V}$\;

\BlankLine
\SetKwFunction{FRecurs}{Decompose}
\SetKwProg{Fn}{Function}{:}{}
\Fn{\FRecurs{$v, d$}}{
    \If{$d < D$}{
        \tcp{LLM generates subcategories}
        $\mathcal{C}_v \leftarrow \textsc{Expand}_{\mathcal{M}}(v,\delta_v)$\;
        \If{$\mathcal{C}_v \neq \emptyset$}{
            \ForEach{$u \in \mathcal{C}_v$}{
                $\text{parent}(u) \leftarrow v$\;
                $\delta_u \leftarrow \textsc{Describe}_{\mathcal{M}}(u)$\;
                $\mathcal{V} \leftarrow \mathcal{V} \cup \{u\}$\;
                \tcp{Recursive DFS call}
                \FRecurs{$u, d+1$}\;
            }
        }
    }
}
\end{algorithm}

\subsection{Algorithm -- Hallucination Injection Template Construction}

Algorithm~\ref{alg:template_construction} details the procedure for generating and filtering sentence-level hallucination templates. The process iterates through each sentence $s$ within the review corpus. To circumvent the intractability of naively instantiating every template, we first apply a coarse-grained screening by evaluating which first-level anchor concepts $c \in \mathbf{V}^{(1)}$ are applicable to the context. For each compatible anchor, fine-grained templates are constructed exclusively from its descending leaf nodes $\mathbf{V}_{\text{leaf}}^{(c)}$, forming the initial candidate set $\mathcal{T}_s^{\text{candidate}}$. Subsequently, we leverage an LLM-based compatibility check $\textsc{Check}_{\mathcal{M}}(T, s)$ guided by a dedicated screening prompt (detailed in Appendix~\ref{app: prompt_template_feasibility_check}). This semantic validation filters out templates whose hallucination types cannot naturally map onto the target sentence, ultimately yielding the feasible template set $\mathcal{T}^{\text{feasible}}$.

\subsection{Algorithm -- Automated Hallucination Injection Pipeline with Semantic Verification}
Algorithm~\ref{alg:hallucination_injection_verified} describes the automated injection pipeline that turns feasible templates into verified hallucinated sentences. For each feasible template $(s, a_s, \delta_v, \pi_v)$, the LLM injector $\textsc{Inject}_{\mathcal{M}}$ rewrites the original sentence $s$ according to the taxonomy instruction $\delta_v$, producing a candidate hallucinated sentence $\tilde{s}$. To ensure that the injection actually altered the meaning, a post-hoc verifier $\textsc{Verify}_{\mathcal{M}}$ compares $\tilde{s}$ against $s$ and returns whether the two are semantically equivalent. A candidate is retained only when $\textsc{Verify}_{\mathcal{M}}(s, \tilde{s}) = 0$, i.e., the generated sentence is judged \emph{not} equivalent to the original and therefore carries a genuine hallucination; equivalent rewrites (output $1$) are discarded as failed injections. The procedure returns the set of verified hallucinated sentences $\{\tilde{s}\}$.







\begin{algorithm}[th]
\footnotesize
\SetInd{0.4em}{0.7em}
\caption{Sentence-level Hallucination Injection Template Construction}
\label{alg:template_construction}

\KwIn{Set of papers $\mathcal{P}$, hierarchical taxonomy node set $\mathbf{V}$ (where $\mathbf{V}^{(1)} \subset \mathbf{V}$ denotes first-level anchors), LLM $\mathcal{M}$ for template compatibility checking}
\KwOut{Sentence-level feasible templates $\{\mathcal{T}^{\text{feasible}}\}$}

\ForEach{$p \in \mathcal{P}$}{
    \ForEach{$r \in \mathcal{R}_p$}{
        \ForEach{$s \in r$}{
            $\mathcal{T}_s^{\text{candidate}} \gets \emptyset$\;
            
            \tcp{Step 1: Coarse-grained screening via anchor nodes from $\mathbf{V}$}
            \ForEach{anchor node $c \in \mathbf{V}^{(1)}$}{
                \If{$\textsc{IsApplicable}(s, c)$}{
                    \tcp{Extract leaf nodes descending from anchor $c$}
                    $\mathbf{V}_{\text{leaf}}^{(c)} \gets \{v \in \mathbf{V} \mid v \text{ is a leaf descendant of } c\}$\;
                    $\mathcal{T}_{\mathcal{L}} \gets \{T_{s,v} \mid v \in \mathbf{V}_{\text{leaf}}^{(c)}\}$\;
                    $\mathcal{T}_s^{\text{candidate}} \gets \mathcal{T}_s^{\text{candidate}} \cup \mathcal{T}_{\mathcal{L}}$\;
                }
            }

            \tcp{Step 2: Compatibility check via LLM screening prompt}
            $\mathcal{T}^{\text{feasible}} \gets \emptyset$\;
            \ForEach{$T \in \mathcal{T}_s^{\text{candidate}}$}{
                \If{$\textsc{Check}_{\mathcal{M}}(T, s) = \text{feasible}$}{
                    $\mathcal{T}^{\text{feasible}} \gets \mathcal{T}^{\text{feasible}} \cup \{T\}$\;
                }
            }
        }
    }
}
\Return $\{\mathcal{T}^{\text{feasible}}\}$\;
\end{algorithm}

\begin{algorithm}[t]
\footnotesize
\SetInd{0.4em}{0.7em}
\caption{Automated Hallucination Injection Pipeline with Semantic Verification}
\label{alg:hallucination_injection_verified}

\KwIn{Sentence-level feasible templates $\{\mathcal{T}^{\text{feasible}}\}$, 
LLM injector $\textsc{Inject}_\mathcal{M}$, LLM $\textsc{Verify}_\mathcal{M}$ for post-hoc semantic verification}

\KwOut{Hallucinated sentences $\{\tilde{s}\}$}

    \ForEach{$(s, a_s, \delta_v, \pi_v) \in \mathcal{T}_s^{\text{feasible}}$}{
        \tcp{Generate hallucinated sentence}
        $\tilde{s} \gets \textsc{Inject}_\mathcal{M}(s, \delta_v)$\;

        \tcp{Post-hoc semantic verification}
        \If{$\textsc{Verify}_\mathcal{M}(s, \tilde{s}) = 0$}{
            \tcp{Retain hallucinated sentence if semantically distinct}
            Store $\tilde{s}$\;
        }
    }

\Return $\{\tilde{s}\}$\;
\end{algorithm}






\begin{table*}[!htbp]
\centering
\footnotesize
\setlength{\tabcolsep}{4pt}
\renewcommand{\arraystretch}{1.25}
\caption{Examples of paper-review hallucinations categorized by taxonomy depth, review content, and verification reasoning.}
\label{tab:hallu-taxonomy-examples}
\begin{tabular}{@{}p{0.24\linewidth} p{0.35\linewidth} p{0.37\linewidth}@{}}
\toprule
\textbf{Class (Depth 1 $\rightarrow$ 2 $\rightarrow$ 3)} & \textbf{Review Content} & \textbf{Labor Reasoning} \\
\midrule

\paperrow{End-to-end Learnable Clustering for Intent Learning in Recommendation}
\textbf{Entity} \newline 
$\hookrightarrow$ Claim Entity Extension \newline 
$\hookrightarrow$ Attribute Substitution
&
The paper introduces Intent-Coupled Contrastive Learning (ICL), a groundbreaking method that significantly enhances user embeddings by incorporating intent information.
&
The paper explicitly names the method ``intent-assisted contrastive learning'' (ICL). The reviewer alters the algorithm's core attribute name (assisted $\rightarrow$ coupled) while retaining its acronym and structure.
\keyquote{Section 3.4.3. Intent-assisted contrastive learning}
\\
\addlinespace[3pt]

\paperrow{Opponent Modeling based on Subgoal Inference}
\textbf{False Concatenation} \newline 
$\hookrightarrow$ Entity Misattribution \newline 
$\hookrightarrow$ Method Misattribution
&
- L204 and L206-207 --- both sentences here claim seemingly contradictory statements: ``...adopting an optimistic strategy akin to the minimax strategy, which applies to cooperative games'' seems to contradict the following statement ``...leading to a conservative strategy similar to the minimax strategy, which is commonly used for general-sum games''.
&
The reviewer misquotes the paper by attributing the ``minimax'' strategy (a method described for general-sum games) to cooperative games, substituting it for the actual method (``maximax''). This methodological misattribution causes the reviewer to incorrectly concatenate the two distinct game contexts and fabricate a false contradiction.
\keyquote{thus adopting an optimistic strategy akin to the maximax strategy [6], which applies to cooperative games}
\\
\addlinespace[3pt]

\paperrow{EZ-HOI: VLM Adaptation via Guided Prompt Learning for Zero-Shot HOI Detection}
\textbf{Number} \newline 
$\hookrightarrow$ Dimensional Parameter Error \newline 
$\hookrightarrow$ Scale Parameter Error
&
``, why is the threshold value chosen to be 0.5 instead of 0.7 or other?
&
The paper explicitly states that the confidence threshold for the object detector is $\theta=0.2$. The reviewer hallucinates the algorithmic scale parameters ($0.5$ and $0.7$) as the premise for their clarification question. \textit{Note:} While interrogative sentences are often excluded from strict hallucination judgments, the foundational premise of the question contains a clear numerical parameter distortion.
\keyquote{We use an off-the-shelf object detector and add a threshold $\theta$ to filter out some low-confident predictions and we set $\theta=0.2$}
\\
\addlinespace[3pt]

\paperrow{ParallelEdits: Efficient Multi-Aspect Text-Driven Image Editing with Attention Grouping}
\textbf{Attribution Failure} \newline 
$\hookrightarrow$ Misattributed Result \newline 
$\hookrightarrow$ Mischaracterized Paper Result
&
Figure 1 illustrates the swapping of multiple objects within the same image, but it only demonstrates the addition of a single object.
&
The reviewer mischaracterizes the scope of the paper's qualitative results by claiming Figure 1 ``only demonstrates the addition of a single object.'' In reality, the evidence explicitly describes Figure 1 as showing multi-aspect edits, including background changes, object removal, and object swapping. The reviewer artificially narrows the scope of the presented results.
\keyquote{such as adding a necktie to a cat and changing the background wall to a beach (Fig.\ 1, Left), or removing a man}
\\
\addlinespace[3pt]

\bottomrule
\end{tabular}
\end{table*}

\begin{table*}[!htbp]
\centering
\footnotesize
\setlength{\tabcolsep}{4pt}
\renewcommand{\arraystretch}{1.25}
\caption{Examples of paper-review hallucinations categorized by taxonomy depth, review content, and verification reasoning.}
\label{tab:hallu-taxonomy-examples-2}
\begin{tabular}{@{}p{0.24\linewidth} p{0.35\linewidth} p{0.37\linewidth}@{}}
\toprule
\textbf{Class (Depth 1 $\rightarrow$ 2 $\rightarrow$ 3)} & \textbf{Review Content} & \textbf{Labor Reasoning} \\
\midrule

\paperrow{Diffusion Models are Certifiably Robust Classifiers}
\textbf{Context Meaning Error} \newline 
$\hookrightarrow$ Jargon Misapplication \newline 
$\hookrightarrow$ Acronym Misinterpretation
&
Then, it proposes Exact Posterior Noised Diffusion Classifier (EPNDC) and Approximated Posterior Noised Diffusion Classifier (APNDC) by deriving ELBO upper bounds on $\log p(x_\tau)$ and thereby enabling classifying noisy images.
&
ELBO stands for Evidence \emph{Lower} Bound, so calling them ``upper bounds'' is wrong. And the paper actually bounds $\log p(x_\tau \mid y)$, not $\log p(x_\tau)$.
\keyquote{we generalize diffusion classifiers to calculate $p(y \mid x_\tau)$ by estimating $\log p(x_\tau \mid y)$ using its ELBO}
\\
\addlinespace[3pt]

\paperrow{Q-VLM: Post-training Quantization for Large Vision-Language Models}
\textbf{Context Meaning Error} \newline 
$\hookrightarrow$ Jargon Misapplication \newline 
$\hookrightarrow$ Jargon Term Substitution
&
The authors separate several layers in a LVLM into blocks and search for the optimal quantization bitwidth for each block individually.
&
The reviewer replaces the paper's actual methodological jargon (``rounding function'') with an incorrect term (``quantization bitwidth''). In the field of model quantization, determining the optimal rounding function is conceptually distinct from determining bitwidth.
\keyquote{Searching the optimal rounding function by considering the output quantization errors for each block achieves better trade-off between the search cost and the quantization accuracy}
\\
\addlinespace[3pt]

\paperrow{Fantasy: Transformer Meets Transformer in Text-to-Image Generation}
\textbf{Entity} \newline 
$\hookrightarrow$ Claim Entity Ext. \newline 
$\hookrightarrow$ Attribute Substitution
&
Unlike commonly used text encoders like CLIP and T5, this study introduces an efficient decoder-only LLM, phi-3, achieving better semantic understanding.
&
The reviewer replaces the specific model entity used in the paper (``Phi-2'') with a different model version (``phi-3''). Although the discrepancy surfaces as a digit change, ``Phi-2'' and ``Phi-3'' represent distinct algorithmic entities in NLP. The reviewer alters the claim's core algorithmic attribute by substituting the named entity.
\keyquote{we employ Phi-2 [24], a state-of-the-art, lightweight LLM, as the text encoder}
\\
\addlinespace[3pt]

\paperrow{Generalization Bound and Learning Methods for Data-Driven Projections in Linear Programming}
\textbf{Reasoning Error} \newline 
$\hookrightarrow$ Mischaracterized Method. \newline 
$\hookrightarrow$ Mischaracterized Algorithm
&
To achieve this, it is necessary to choose a good $P$ that minimizes the empirical optimal value.
&
The reviewer states the methodological goal is to ``minimize'' the empirical optimal value. However, the paper explicitly aims to \emph{maximize} the expected/empirical optimal value via gradient ascent. By completely reversing the direction of optimization, the reviewer fundamentally mischaracterizes the paper's core computational technique and algorithm.
\keyquote{we can use the gradient ascent method to maximize $u(P, \pi)$ under the regularity condition}
\\
\addlinespace[3pt]

\paperrow{DataStealing: Steal Data from Diffusion Models in Federated Learning with Multiple Trojans}
\textbf{Overgeneralization} \newline 
$\hookrightarrow$ Result-Based Overgen. \newline 
$\hookrightarrow$ Domain Extrapolation
&
- \textbf{Broad Applicability}: The proposed methodologies and findings are not limited to a specific application but are broadly applicable to various domains where FL and generative models are used.
&
The reviewer claims the paper's findings are broadly applicable to various domains involving federated learning (FL) and generative models. However, the paper specifically targets and tests only diffusion models in FL for image generation (e.g., CIFAR10, CelebA). The reviewer unjustifiably extrapolates a narrow, diffusion-specific empirical result to all FL and generative-model domains without any experimental evidence or justification provided in the paper.
\keyquote{We propose an attack method for DataStealing, named AdaSCP, to defeat advanced distance-based defenses and seamlessly incorporate backdoor gradients into the global diffusion model.}
\\

\bottomrule
\end{tabular}
\end{table*}

\begin{table*}[!htbp]
\centering
\footnotesize
\setlength{\tabcolsep}{4pt}
\renewcommand{\arraystretch}{1.25}
\caption{Examples of paper-review hallucinations categorized by taxonomy depth, review content, and verification reasoning.}
\label{tab:hallu-taxonomy-examples-3}
\begin{tabular}{@{}p{0.24\linewidth} p{0.35\linewidth} p{0.37\linewidth}@{}}
\toprule
\textbf{Class (Depth 1 $\rightarrow$ 2 $\rightarrow$ 3)} & \textbf{Review Content} & \textbf{Labor Reasoning} \\
\midrule
\paperrow{Convolutional Differentiable Logic Gate Networks}
\textbf{Hyperbole} \newline 
$\hookrightarrow$ Inflated Comparison Advantage \newline 
$\hookrightarrow$ Baseline Performance Fabrication
&
Lowest latency of all SOTA baseline results, the majority of them being much slower with even worse accuracy.
&
While the latency claim is well supported, the accuracy comparison is highly exaggerated. The reviewer asserts that the majority of baselines have ``even worse accuracy'' than the proposed LogicTreeNet-B. However, Table 1 shows this baseline performance claim is inconsistent with the data; many baselines (e.g., BinaryNet, FBNA CNV) actually achieve \textit{higher} accuracy. The reviewer fabricates an inconsistent baseline trend to inflate the paper's comparative advantage beyond what the data shows.
\keyquote{LogicTreeNet-B 80.17\% 24 ns ... BinaryNet [29] 88.60\% 4 090 M Zhao et al. [30] 88.54\% 4 940 M FBNA CNV [31] 88.61\% 5 540 M}
\\

\paperrow{Self-Guided Masked Autoencoders for Domain-Agnostic Self-Supervised Learning}
\textbf{Hyperbole} \newline 
$\hookrightarrow$ Inflated Comparative Advantage \newline 
$\hookrightarrow$ Mischaracterized Comparative Setting
&
On all evaluated benchmarks, SMA not only competes but surpasses the state-of-the-art, indicating its potential as a leading approach in self-supervised learning.
&
The review overstates the scope of the paper's SOTA claim. The abstract limits the claim to \textit{three} benchmarks---protein biology, chemistry, and particle physics---stating SMA ``achieves state-of-the-art performance on these three benchmarks.'' The review instead asserts superiority over the state-of-the-art on \textit{all} evaluated benchmarks. This is not supported: other evaluated settings, such as GLUE in Table~4, compare SMA only against internal baselines (No Pretrain, Random, and Word-masking) rather than the state-of-the-art. By generalizing a SOTA claim scoped to three benchmarks into a universal one, the reviewer mischaracterizes the comparative setting.
\keyquote{achieves state-of-the-art performance on these three benchmarks}
\\

\bottomrule
\end{tabular}
\end{table*}

\section{Prompt Example}
\label{app:prompt_example}
\subsection{Prompt -- Hallucination Taxonomy Generation}
\label{app: prompt_hallucination_taxonomy_generation}
Fig.~\ref{fig:prompt_hallucination_taxonomy_generation} presents the prompt template used for automatic hallucination taxonomy generation in our experiment.

\subsection{Prompt -- Hallucination Taxonomy Overlap Filtering}
\label{app: prompt_hallucination_taxonomy_filter}
Fig.~\ref{fig:prompt_similar_filter} presents the prompt template used for automatic hallucination taxonomy overlap filtering in our experiment.

\subsection{Prompt -- Coarse Category Screening}
\label{app: prefilter}
Fig.~\ref{fig:prompt_coarse_screening} presents the prompt template used for coarse category screening of taxonomy.

\subsection{Prompt -- Template Construction}
\label{app: template_construction}
Fig.~\ref{fig:prompt_template_construction} presents the prompt template used for template construction.

\subsection{Prompt -- Review Filtering}
\label{app: prompt_review_filtering}
Fig.~\ref{fig:prompt_review_filtering} presents the prompt template used for review filtering in our experiment.

\subsection{Prompt -- Review Aspect Tagging}
\label{app: prompt_review_aspect_tagging}
Fig.~\ref{fig:prompt_review_aspect_tagging} presents the prompt template used for automatic review aspect tagging in our experiment.

\subsection{Prompt -- Hallucination Template Feasibility Check}
\label{app: prompt_template_feasibility_check}
Fig.~\ref{fig:prompt_template_feasibility_check} presents the prompt template used for the automatic hallucination template feasibility check in our experiment.

\subsection{Prompt -- Hallucination Template Injection}
\label{app: prompt_template_injection}
Fig.~\ref{fig:prompt_template_injection} presents the prompt template used for the automatic hallucination template injection in our experiment.

\subsection{Prompt -- Post-hoc Verification}
\label{app: prompt_posthoc_verification}
Fig.~\ref{fig:prompt_posthoc_verification} presents the prompt template used for the automatic post-hoc verification in our experiment.

\subsection{Prompt -- RA-LLM (KR)}
\label{app: prompt_RA-LLM_KR}
Fig.~\ref{fig:prompt_RA-LLM_KR} presents the prompt template used for the RA-LLM (KR) baseline in task 1 evaluation (Hallucination Detection).

\subsection{Prompt -- RA-LLM (CoT)}
\label{app: prompt_RA-LLM_CoT}
Fig.~\ref{fig:prompt_RA-LLM_CoT} presents the prompt template used for the RA-LLM (CoT) baseline in task 1 evaluation (Hallucination Detection).

\subsection{Prompt -- RA-LLM (Contrast)}
\label{app: prompt_RA-LLM_contrast}
Fig.~\ref{fig:prompt_RA-LLM_Contrast} presents the prompt template used for the RA-LLM (Contrast) baseline in task 1 evaluation (Hallucination Detection).

\subsection{Prompt -- General-Purpose LLM for Task 1} Fig.~\ref{fig:prompt_hallucination_detection} presents the prompt template used for the  General-Purpose LLM in task 1 evaluation (Hallucination Detection).


\subsection{Prompt -- General-Purpose LLM for Task 2}
Fig.~\ref{fig:prompt_hallucination_classification} presents the prompt template used for the  General-Purpose LLM in task 2 evaluation (Hallucination Classification).


\subsection{Prompt -- General-Purpose LLM for Task 3}
Fig.~\ref{fig:prompt_hallucination_selection} presents the prompt template used for the  General-Purpose LLM in task 3 evaluation (Hallucination Span Localization).


\begin{figure*}
\centering
\begin{tcolorbox}[
  colback=gray!5,
  colframe=gray!180,
  title=Core Judgement Criteria for Hallucination Taxonomy Refinement,
  fonttitle=\bfseries,
  fontupper=\small,
  boxrule=0.5pt,
  arc=2mm,
  left=2mm, right=2mm, top=1mm, bottom=1mm,
  width=\textwidth
]
 Your task is to assess candidate concept nodes against the following core judgement criteria to ensure the taxonomy is analytically rigorous, MECE-compliant, and operationally reliable for human annotation.
\\\\
\textbf{\#\# 1. Global Architecture Uniqueness \& Redundancy Minimization}\\
Each hallucination type must have a unique and unambiguous position in the taxonomy tree.\\
- \textbf{Path Exclusivity}: If a concept is already defined as an independent item at the first level (Depth 1), it must not be duplicated under other categories (\textit{e.g.}, Reasoning Error).\\
- \textbf{Annotation Conflict Prevention}: Avoid forcing annotators to choose between two logical paths when confronting the same error, as this reduces data consistency.\\
- \textbf{Redundancy Minimization}: The taxonomic structure must remain concise; structural over-stacking is strictly prohibited.\\
- \textbf{Synonym Merging}: Nodes with highly overlapping definitions or that are merely near-synonyms must be merged (\textit{e.g.}, Citation Year Substitution and Publication Year Inaccuracy).\\
- \textbf{Structural Flattening}: If a child node's definition is fully subsumed by its parent and provides no additional operational value, it must be simplified.
\\\\
\textbf{\#\# 2. Parent-Child Alignment and Logical Inheritance}\\
\textbf{Evaluation Standard}: Does the node logically constitute a proper subset of its parent node?\\
- \textbf{Logical Subordination}: Child nodes must precisely inherit the categorical scope of their parent node.\\
- \textbf{Retention Principle}: Prioritize options that best fit the parent concept.\\
- \textbf{Elimination Mechanism}: If a node's definition exceeds the parent node's scope or is misclassified, it must be eliminated or restructured, regardless of definitional quality.
\\\\
\textbf{\#\# 3. Conceptual Precision and Actionability}\\
\textbf{Evaluation Standard}: Is the node's definition concrete enough for annotators (or models) to identify unambiguously when encountering actual data?\\
- \textbf{Retention Principle}: Prioritize nodes that provide concrete mechanisms, clear boundaries, or specific examples (\textit{e.g.}, particular lexical cues, specific comparison targets).\\
- \textbf{Elimination Mechanism}: Eliminate overly general, vaguely worded, or redundant nodes that merely restate the parent definition.
\\\\
\textbf{\#\# 4. Coverage and Mutual Exclusivity (MECE Principle)}\\
\textbf{Evaluation Standard}: Are two nodes fully overlapping (synonymous), partially overlapping, or mutually exclusive?\\
- \textbf{Near-Synonymous Case}: Select the node with more rigorous wording and stronger alignment with academic terminology.\\
- \textbf{Containment Relation}: Prioritize the node that fills a missing level in the current taxonomy tree; if two nodes share the same depth but differ in scope, retain the one with the more precise definition.\\
- \textbf{Fully Distinct Mechanisms (Mutually Exclusive)}: Retain both nodes.
\\\\
\textbf{\#\# 5. Nomenclature and Structural Consistency}\\
\textbf{Evaluation Standard}: Does the node name conform to the naming conventions of the overall taxonomy (\textit{e.g.}, uniform adoption of the ``Phenomenon + via + Mechanism'' format)?\\
- \textbf{Structural Alignment}: Select options whose naming style aligns with other nodes in the system, especially those at the same hierarchical level.
\end{tcolorbox}
\caption{Core judgement criteria used for hallucination taxonomy refinement.}
\label{fig:criteria_hallucination_taxonomy_refinement}
\end{figure*}

\begin{figure*}
\centering
\begin{tcolorbox}[
  colback=gray!5,
  colframe=gray!180,
  title=Prompt Template for Hallucination Taxonomy Generation,
  fonttitle=\bfseries,
  fontupper=\small, 
  boxrule=0.5pt,
  arc=2mm,
  left=2mm, right=2mm, top=1mm, bottom=1mm,
  width=\textwidth 
]
You are an expert in Large Language Model (LLM) hallucination taxonomy construction for scientific paper reviews. Your task is to decompose a given parent hallucination concept into operationally distinguishable child concepts at a specified taxonomy depth. The goal is to construct a taxonomy that is analytically rigorous, MECE-compliant, and suitable as an annotation label space. The taxonomy must support reliable human annotation and reviewer agreement. A critical requirement is that concept boundaries are clear and enforceable based on observable review text, not abstract error theory.
\\\\
\textbf{\#\# TASK OVERVIEW}\\
Given a parent hallucination concept, generate child concepts that represent concrete hallucination realization patterns at the current taxonomy depth.
 
Each child concept must be defined by:\\
- a concise and stable concept name (used as the annotation label), and\\
- an operational concept description that explains how the hallucination manifests in review text.
 
The focus is strictly on *textually observable hallucination patterns in reviews*, not on model intent, generation causes, or abstract error classes.
\\\\
\textbf{\#\# INPUT SLOTS}\\
- Ancestor Path (from root to parent, ordered from most general to most specific):\\
Use this to calibrate how much narrowing each level represents, and match that pace for the children you generate.\\
\{ancestor\_path\}
 
- Parent Concept (Dictionary):\\
\{"concept": \{parent\_concept\}, "description": \{parent\_concept\_desc\}\}
 
- Current Taxonomy Depth:
\{current\_depth\}
 
- Existing Sibling Concepts at This Depth (List of Dictionaries, for MECE checking):\\
Each sibling concept is provided as:
[
  \{
  "concept": "...", 
  "description": "..."
  \}, 
  ...
]\\
\{sibling\_concepts\}
 
- Maximum Number of Child Concepts to Generate:
\{max\_width\}
\\\\
\textbf{\#\# GENERATION CONSTRAINTS}\\
1. MECE Compliance\\  
- Each generated child concept must be mutually exclusive with all provided sibling concepts at this depth.\\
- The set of generated child concepts must collectively exhaust the parent concept.\\
- Distinctions must be justified at the level of observable textual evidence in review text.
 
2. Operational Specificity\\  
Each child concept description must explicitly specify:\\
- What aspect of the paper is hallucinated (\textit{e.g.}, method, result, experiment, motivation, comparison, conclusion).\\
- How it is hallucinated (\textit{e.g.}, fabrication, attribute substitution, unsupported extension, semantic distortion).\\
- Typical linguistic cues or claim structures in the review text that indicate this hallucination.
 
3. Depth Awareness\\  
- Use current depth AND ancestor path together to determine abstraction level.\\
\hspace*{1em}- Shallow depths (depth 2): generate abstract categorical distinctions.\\
\hspace*{1em}- Deeper depths (depth 3+): permit concrete, instance-level distinctions only when the parent is already an appropriately narrow category.\\
- Do NOT jump to instance-level specifics at intermediate depths. If concrete variants exist, first define the abstract category that contains them; defer the instances to the next level down.\\
- Child concepts should be parallel in granularity to existing siblings.
 
4. Minimality\\ 
- Generate only the minimum number of child concepts required to fully cover the parent concept.\\
- Avoid stylistic variants or overlapping distinctions that are not operationally enforceable.
\\\\
\textbf{\#\# OUTPUT FORMAT}\\
The output MUST be a single JSON array.
 
If child concepts are generated, return an array of objects, each representing one immediate child concept:
 
[
  \{
    "concept": "<canonical label>",
    "description": "<operational definition based on observable review text>"
  \}
]
 
Field constraints:\\
- concept:\\
\hspace*{1em}- Short, stable, canonical label\\
\hspace*{1em}- Noun phrase only\\
\hspace*{1em}- No punctuation, qualifiers, or examples
 
- description:\\
\hspace*{1em}- Single sentence only (no lists, no commentary, no causal speculation)\\
\hspace*{1em}- Operational definition grounded in textually observable review content\\
\hspace*{1em}- Specifies what aspect of the paper is hallucinated and how it manifests in review text
 
If the parent concept is atomic at the current taxonomy depth, return:
 
[]
\end{tcolorbox}
\caption{Prompt template used for hallucination taxonomy generation.}
\label{fig:prompt_hallucination_taxonomy_generation}
\end{figure*}

\begin{figure*}
\centering
\begin{tcolorbox}[
  colback=gray!5,
  colframe=gray!180,
  title=Prompt Template for Near-Synonym Filtering,
  fonttitle=\bfseries,
  fontupper=\small, 
  boxrule=0.5pt,
  arc=2mm,
  left=2mm, right=2mm, top=1mm, bottom=1mm,
  width=\textwidth 
]
You are an expert in semantic taxonomy alignment.
 
Your task is to determine whether two taxonomy nodes are near-synonyms, considering:\\
- concept meaning\\
- concept descriptions\\
- hierarchical context (ancestor path)\\
- level of abstraction (depth)
 
Two nodes are considered "near-synonyms" ONLY IF:\\
- They refer to the same or almost identical concept in this taxonomy context\\
- They would be redundant if both kept in the same tree\\
- Differences in wording do NOT imply different scope or abstraction
---
\textbf{\#\#\# Node A}\\
- Concept: \{concept\_A\}\\
- Description: \{description\_A\}\\
- Ancestor Path (root $\rightarrow$ node): \{path\_A\}\\
- Depth: \{depth\_A\}
 
\textbf{\#\#\# Node B}\\
- Concept: \{concept\_B\}\\
- Description: \{description\_B\}\\
- Ancestor Path (root $\rightarrow$ node): \{path\_B\}\\
- Depth: \{depth\_B\}
---
\textbf{\#\#\# Instructions}\\
Step 1: Compare semantic meaning ignoring wording differences.
 
Step 2: Compare taxonomy context:\\
- Are they under similar parent concepts?\\
- Do they represent the same level of abstraction?\\
- Would merging them break the taxonomy structure?
 
Step 3: Decide:
 
Output in JSON format:
 
\{
  "is\_synonym": True/False,
  "keep": "A" | "B" | "NA",
  "explanation": "string or NA"
\}
 
Rules:\\
- If False:\\
\hspace*{1em}- keep = "NA"\\
\hspace*{1em}- explanation = "NA"
 
- If True:\\
\hspace*{1em}- Choose the node that is:\\
\hspace*{2em}* more precise\\
\hspace*{2em}* better aligned with its ancestor path\\
\hspace*{2em}* more consistent with taxonomy naming style\\
\hspace*{1em}- explanation must include:\\
\hspace*{2em}* why they are synonyms\\
\hspace*{2em}* why the chosen node is better in context
 
Be strict: do NOT mark as synonyms if there is any meaningful difference in scope, abstraction level, or taxonomy role.
\end{tcolorbox}
\caption{Prompt template used for near-synonym hallucination taxonomy filtering.}
\label{fig:prompt_similar_filter}
\end{figure*}

\begin{figure*}
\centering
\begin{tcolorbox}[
  colback=gray!5,
  colframe=gray!180,
  title=Prompt Template for Review Filtering,
  fonttitle=\bfseries,
  fontupper=\small, 
  boxrule=0.5pt,
  arc=2mm,
  left=2mm, right=2mm, top=1mm, bottom=1mm,
  width=\textwidth 
]
You are an expert in academic peer-review analysis. Your task is to determine whether the meta review substantively adopted the core opinion of a specific reviewer. You will be given information about ONE reviewer only. Do NOT consider other reviewers except as reflected in the meta review.

---

\textbf{\#\#\# Input}

[Reviewer Review]
\{reviewer\_review\}

[Meta Review]
\{meta\_review\}

---

\textbf{\#\#\# Instructions}

1. Identify the reviewer's core opinion, i.e., the main reasons for acceptance or rejection.\\
2. Examine whether the meta review:\\
\hspace*{1em}- explicitly endorses or echoes these points, OR\\
\hspace*{1em}- implicitly relies on them in its reasoning, OR\\
\hspace*{1em}- explicitly rejects or ignores them.\\
3. If the meta review adopts the same reasoning even without mentioning the reviewer explicitly, this still counts as adoption.

---

\textbf{\#\#\# Output Format (strictly follow)}

Return a valid JSON object ONLY, using EXACTLY the following template.
You must fill in the values but MUST NOT change any key names.

\{
  "adopted": <true | false>,
  "confidence": "<high | medium | low>",
  "justification": "<one concise sentence explaining how the meta review does or does not adopt the reviewer's core opinion.>"
\}

---

\textbf{\#\#\# Decision Criteria}

- adopted = true:
  The meta review clearly or implicitly incorporates the reviewer’s core arguments
  as part of its justification.

- adopted = false:
  The meta review explicitly contradicts the reviewer’s core arguments,
  ignores their main concerns, or reaches a decision based on different reasoning.

- If the evidence is indirect or ambiguous, lower the confidence accordingly.
\end{tcolorbox}
\caption{Prompt template used for review filtering.}
\label{fig:prompt_review_filtering}
\end{figure*}

\begin{figure*}
\centering
\begin{tcolorbox}[
  breakable,
  colback=gray!5,
  colframe=gray!180,
  title=Prompt Template for Coarse-grained Feasibility Screening,
  fonttitle=\bfseries,
  fontupper=\footnotesize,
  boxrule=0.5pt,
  arc=2mm,
  left=2mm, right=2mm, top=1mm, bottom=1mm,
  width=\textwidth
]
\textbf{\#\# ROLE}\\
You are a hallucination-injection feasibility screener. Given a sentence from a scientific paper review, determine which hallucination categories have the \textbf{required raw material} present in the sentence for injection to be possible.

\medskip
\textbf{\#\# TASK}\\
For each category below, check whether the sentence contains the specific linguistic or semantic features listed as prerequisites. Select a category \textbf{only if its prerequisites are clearly satisfied}. When in doubt, exclude.

\medskip
\textbf{\#\# CATEGORIES AND PREREQUISITES}

\smallskip
\textbf{Number}\\
Prerequisite: The sentence contains at least one explicit numeric value --- including percentages, ratios, measurements, counts, years, dimensions, or quantitative comparisons (e.g., ``3 layers'', ``92.4\% accuracy'', ``2018'', ``4x faster'').\\
Exclude if: the sentence only mentions quantity implicitly or qualitatively (e.g., ``several'', ``many'', ``large'').

\smallskip
\textit{$\cdots$ (Entity, False Concatenation, Attribution Failure, Overgeneralization, Reasoning Error, Hyperbole, Temporal omitted) $\cdots$}

\smallskip
\textbf{Context-based meaning error}\\
Prerequisite: The sentence contains a term, abbreviation, or phrase that has \textbf{multiple plausible interpretations} in a scientific context --- including overloaded acronyms, domain-specific terms with general-language counterparts, or phrasing whose meaning depends on surrounding context.\\
Exclude if: all terms in the sentence have a single unambiguous meaning in context.

\medskip
\textbf{\#\# INPUT}\\
Injection\_Sentence: ``\{injection\_sentence\}''

\medskip
\textbf{\#\# OUTPUT FORMAT (STRICT)}\\
Output ONLY a JSON array of the category names whose prerequisites are satisfied. No explanations. No extra text.

\smallskip
Valid names: ``Number'', ``Entity'', ``False Concatenation'', ``Attribution Failure'', ``Overgeneralization'', ``Reasoning Error'', ``Hyperbole'', ``Temporal'', ``Context-based meaning error''

\smallskip
Example:\\
{[}``Number'', ``Hyperbole'', ``Reasoning Error''{]}
\end{tcolorbox}
\caption{Prompt template used for coarse-grained feasibility screening before fine-grained template construction. Seven of the nine categories are omitted for brevity; all nine follow the same prerequisite/exclude format.}
\label{fig:prompt_coarse_screening}
\end{figure*}

\begin{figure*}
\centering
\begin{tcolorbox}[
  breakable,
  colback=gray!5,
  colframe=gray!180,
  title=Prompt Template for Injection Template Construction,
  fonttitle=\bfseries,
  fontupper=\footnotesize,
  boxrule=0.5pt,
  arc=2mm,
  left=2mm, right=2mm, top=1mm, bottom=1mm,
  width=\textwidth
]
You are a compatibility checker for hallucination injection templates in scientific paper reviews.

\medskip
\textbf{\#\# TASK}\\
Given a template with four fields, decide whether the hallucination defined by \texttt{Hallucination\_Instruction} and \texttt{Hallucination\_Labels} can be coherently injected into \texttt{Injection\_Sentence} while remaining grounded in \texttt{Injection\_Aspect}. Output \texttt{true} only if ALL conditions below hold. Be strict and conservative.

\medskip
\textbf{\#\# COMPATIBILITY CONDITIONS}

\smallskip
1. \textbf{Operational applicability}\\
The operation described in \texttt{Hallucination\_Instruction} can be concretely performed on \texttt{Injection\_Sentence} --- there must exist a target in the sentence to modify, replace, exaggerate, or fabricate.

\smallskip
2. \textbf{Semantic prerequisites of the leaf type}\\
\texttt{Hallucination\_Labels} is a taxonomy path (root > ... > leaf). The LEAF node defines the specific hallucination type. Its semantic prerequisites must be present in \texttt{Injection\_Sentence}.\\
Examples:\\
\hspace*{1em}- Number leaf $\rightarrow$ numeric value present\\
\hspace*{1em}- Entity leaf $\rightarrow$ named entity that can be swapped\\
\hspace*{1em}- Temporal leaf $\rightarrow$ tense / modality / time marker\\
\hspace*{1em}- Reasoning leaf $\rightarrow$ inferential or causal step\\
If the prerequisite is absent, output \texttt{false}.

\smallskip
3. \textbf{Aspect grounding}\\
Every tag in \texttt{Injection\_Aspect} must be substantively supported by \texttt{Injection\_Sentence}. An aspect that the sentence does not actually discuss makes the template incompatible.

\smallskip
4. \textbf{Realistic edit}\\
Applying the hallucination must yield a plausible review sentence, not a nonsensical or category-error edit. If the only way to inject the hallucination is to invent unrelated content, output \texttt{false}.

\medskip
\textbf{\#\# INPUT}\\
Hallucination\_Instruction:\\
\{hallucination\_instruction\}

\smallskip
Injection\_Sentence:\\
\{injection\_sentence\}

\smallskip
Hallucination\_Labels:\\
\{hallucination\_labels\}

\smallskip
Injection\_Aspect:\\
\{injection\_aspect\}

\medskip
\textbf{\#\# OUTPUT (STRICT)}\\
Output exactly one word: \texttt{true} or \texttt{false}. No explanation, no punctuation, no extra text.
\end{tcolorbox}
\caption{Prompt template used for injection template construction.}
\label{fig:prompt_template_construction}
\end{figure*}

\begin{figure*}
\centering
\begin{tcolorbox}[
  colback=gray!5,
  colframe=gray!180,
  title=Prompt Template for Review Aspect Tagging,
  fonttitle=\bfseries,
  fontupper=\small, 
  boxrule=0.5pt,
  arc=2mm,
  left=2mm, right=2mm, top=1mm, bottom=1mm,
  width=\textwidth 
]
You are an expert at aspect-based text classification.
\\\\
\textbf{TASK}:\\
Given a review sentence and a list of candidate aspect labels, identify the SINGLE aspect label that best matches the main focus of the review.
\\\\
\textbf{REVIEW SENTENCE}:
"\{review\_sent\}"

ASPECT LABEL SET (fixed and closed):
\{aspect\_label\_set\}
\\\\
\textbf{INSTRUCTIONS}:\\
- Choose exactly ONE aspect from the candidate list.\\
- Select the aspect that is most semantically relevant to the review sentence.\\
- If multiple aspects seem related, choose the one that is the most central or dominant.\\
- Do NOT create new aspects.\\
- Do NOT explain your reasoning.
\\\\
\textbf{OUTPUT FORMAT (STRICT)}:\\
Return a valid JSON dictionary exactly in the following structure:\\
\{"Aspects": ["Aspect1", "Aspect2", ...]\}
\\
If no aspect applies:
\{"Aspects": ["-"]\}
\\\\
\textbf{FORMAT CONSTRAINTS}:\\
- Output must be valid JSON.\\
- The key must be exactly "Aspects".\\
- Aspect names must exactly match the predefined labels (case-sensitive).\\
- Do not include trailing commas, comments, or extra fields.\\
- Do not include any text outside the JSON output.
\\\\
\textbf{FEW-SHOT EXAMPLES}:
\\\\
REVIEW SENTENCE: The results presented in the paper are significant.\\
OUTPUT: {{"Aspects": ["Significance", "Result"]}}
\\\\
REVIEW SENTENCE: However, authors do not compare their methods against any of the previous works.\\
OUTPUT: \{"Aspects": ["Comparison", "Related Work"]\}
\\\\
REVIEW SENTENCE: The latter is efficient, but the former enables interpretability.\\
OUTPUT: \{"Aspects": ["Efficiency", "Interpretation"]\}
\\\\
REVIEW SENTENCE: In general, I find Experiments II to be much weaker than Experiment I.\\
OUTPUT: \{"Aspects": ["Comparison", "Experiment"]\}
\\\\
REVIEW SENTENCE: (2) The paper does not explain why end-to-end training in the entropy-regularization is necessary.\\
OUTPUT: \{"Aspects": ["Training", "Explanation"]\}
\\\\
REVIEW SENTENCE: Finally, the paper shows that compositional languages generalize better to the held-out validation set.\\
OUTPUT: {{"Aspects": ["Generalization", "Findings"]}}
\\\\
REVIEW SENTENCE: Typo: - Figure 4 caption: "Our variable-resolution inputs prevent**s**..."\\
OUTPUT: \{"Aspects": ["Typo"]\}
\\\\
REVIEW SENTENCE: In the more practical setting of finetuning the model, the attacks are not effective.
OUTPUT: \{"Aspects": ["Fine-tuning", "Effectiveness"]\}
\\\\
REVIEW SENTENCE: The evaluation metric is problematic (or at least unclear).\\
OUTPUT: \{"Aspects": ["Evaluation", "Clarity", "Metric"]\}
\\\\
REVIEW SENTENCE: Overall, this work contributes an interesting framework for analysis.\\
OUTPUT: \{"Aspects": ["Analysis", "Framework"]\}
\\\\
REVIEW SENTENCE: I'm confused as to why some of the visualizations in Fig 3 show white bands along the diagonal.\\
OUTPUT: \{"Aspects": ["Figure", "Confusion"]\}
\\\\
The system must strictly follow the specified output format under all circumstances.
\end{tcolorbox}
\caption{Prompt template used for review aspect tagging.}
\label{fig:prompt_review_aspect_tagging}
\end{figure*}

\begin{figure*}
\centering
\begin{tcolorbox}[
  colback=gray!5,
  colframe=gray!180,
  title=Prompt Template for Hallucination Template Feasibility Check,
  fonttitle=\bfseries,
  fontupper=\small, 
  boxrule=0.5pt,
  arc=2mm,
  left=2mm, right=2mm, top=1mm, bottom=1mm,
  width=\textwidth 
]
You are an automatic compatibility checker for hallucination injection templates.

---

\textbf{\#\# TASK\_OVERVIEW}\\
Your task is to determine whether a given injection template is COMPATIBLE.\\
Be strict and conservative. If the hallucination type cannot be meaningfully applied to the sentence, mark it as incompatible.

A template is COMPATIBLE only if:\\
- The hallucination instruction can be operationally applied to the sentence.\\
- The semantic requirements implied by the hallucination labels are present in the sentence.\\
- The injection aspects are supported by the sentence content.\\
- Applying the hallucination would result in a coherent and realistic edit, not a nonsensical or impossible one.

---

\textbf{\#\# DECISION\_GUIDELINES}\\
- If the hallucination type requires specific properties (\textit{\textit{e.g.}}, numerical values, concrete entities, temporal information, comparisons, or other semantic constraints), those properties must be present in the sentence.\\
- Number-related hallucinations are incompatible with non-quantitative sentences.\\
- Entity-related hallucinations are incompatible if no concrete entity is mentioned.\\
- Hallucination\_Labels are provided as a single taxonomy path (coarse to fine); the leaf node defines the hallucination type and its semantic requirements.\\
- Aspect tags must be grounded in the sentence; unsupported aspects make the template incompatible.\\
- Hallucination\_Labels and Injection\_Aspect define HARD constraints; if either is not satisfied by the sentence, the template is incompatible.\\
- If applying the hallucination would introduce unsupported information or category errors, the template is incompatible.

---

\textbf{\#\# INPUT\_FORMAT}\\
Hallucination\_Instruction:
\{hallucination\_instruction\}

Injection\_Sentence:
\{injection\_sentence\}

Hallucination\_Labels:
\{hallucination\_labels\}

Injection\_Aspect:
\{injection\_aspect\}

---

\textbf{\#\#\# OUTPUT\_FORMAT (STRICT)}\\
You must output exactly ONE token, either:

true
or
false
\\\\
No other output is allowed.
Schema correctness is more important than language quality.
\end{tcolorbox}
\caption{Prompt template used for Hallucination Template Feasibility Check before template injection.}
\label{fig:prompt_template_feasibility_check}
\end{figure*}

\begin{figure*}
\centering
\begin{tcolorbox}[
  colback=gray!5,
  colframe=gray!180,
  title=Prompt Template for Hallucination Template Injection,
  fonttitle=\bfseries,
  fontupper=\small, 
  boxrule=0.5pt,
  arc=2mm,
  left=2mm, right=2mm, top=1mm, bottom=1mm,
  width=\textwidth 
]
You are an expert hallucination generator for review sentences.

---

\textbf{\#\# TASK}\\
Your task is to introduce the specific hallucination described in the instruction into the given sentence.
Follow the instruction precisely and produce a natural, coherent sentence that reflects the hallucination.
Do not alter other parts of the sentence unnecessarily.

---

\textbf{\#\# INPUT\_FORMAT}\\
Hallucination\_Instruction:
\{hallucination\_instruction\}

Injection\_Sentence:
\{injection\_sentence\}

---

\textbf{\#\# OUTPUT\_FORMAT}\\
Return ONLY the modified sentence as a single line of text.\\
Do not add explanations, comments, or JSON.
\end{tcolorbox}
\caption{Prompt template used for Hallucination Template Injection.}
\label{fig:prompt_template_injection}
\end{figure*}

\begin{figure*}
\centering
\begin{tcolorbox}[
  colback=gray!5,
  colframe=gray!180,
  title=Prompt Template for Post-hoc Verification,
  fonttitle=\bfseries,
  fontupper=\small, 
  boxrule=0.5pt,
  arc=2mm,
  left=2mm, right=2mm, top=1mm, bottom=1mm,
  width=\textwidth 
]
\textbf{\#\# TASK}:

You are an expert reviewer specializing in fine-grained semantic comparison of review statements.
Your task is to determine whether two review sentences are strictly semantically equivalent. Equivalence (score = 1) should be rare and assigned only in near-identical paraphrase cases.
In ambiguous cases, choose 0.
\\\\
\textbf{\#\# INPUT}:

Review Sentence A:
\{REVIEW\_SENTENCE\_1\}

Review Sentence B:
\{REVIEW\_SENTENCE\_2\}
\\\\
\textbf{\#\# CRITERIA}:

The two sentences are semantically equivalent ONLY IF ALL of the following are true:\\
- Each sentence fully entails the other.\\
- No opinions, facts, or sentiments are added or omitted.\\
- No difference in emphasis, scope, certainty, or evaluative strength.\\
- No additional qualifiers or implications are introduced.\\
- The sentences are fully interchangeable in the same review context.

If ANY difference exists, the equivalence score MUST be 0.
\\\\
\textbf{\#\# OUTPUT}:

Return the equivalence score as a single digit:\\
- 1: Strict semantic equivalence\\
- 0: Not strictly semantically equivalent

Return only: 0 or 1
\end{tcolorbox}
\caption{Prompt template used for Post-hoc Verification after hallucination template injection.}
\label{fig:prompt_posthoc_verification}
\end{figure*}

\begin{figure*}
\centering
\begin{tcolorbox}[
  colback=gray!5,
  colframe=gray!180,
  title=Prompt Template for RA-LLM (KR),
  fonttitle=\bfseries,
  fontupper=\small, 
  boxrule=0.5pt,
  arc=2mm,
  left=2mm, right=2mm, top=1mm, bottom=1mm,
  width=\textwidth 
]
You are a careful hallucination detector. Answer MUST be Yes or No.
\\[10pt]

\textbf{\#\#\# Instruction} \\
I want you act as an answer judge. Given a question and an answer, your objective is to determine if the provided answer contains non-factual or hallucinated information. You SHOULD give your judgement based on the following hallucination types and the world knowledge.
\\[5pt]

\textit{(Few-shot examples including: misunderstanding context, factual contradiction, specificity level, and logical inference...)}
\\[5pt]


The answer you give MUST be "Yes" or "No".
\\[10pt]
---
\\[10pt]
\textbf{\#\#\# Output Format Requirement} \\
1) You may provide brief reasoning. \\
2) The last line must be exactly: \#Your Judgement\#: Yes or \#Your Judgement\#: No \\
3) Do not output both labels in the final line.
\\[10pt]
---
\\[10pt]
\textbf{\#\#\# Input} \\
\#Question\#: Use the following evidence as knowledge when judging. \\
\{knowledge\} \\
Review statement: \{statement\} \\
\#Answer\#: \{statement\} \\
\#Your Judgement\#:
\end{tcolorbox}
\caption{Prompt template used for RA-LLM (KR) in Task 1 Evaluation.}
\label{fig:prompt_RA-LLM_KR}
\end{figure*}

\begin{figure*}
\centering
\begin{tcolorbox}[
  colback=gray!5,
  colframe=gray!180,
  title=Prompt Template for RA-LLM (CoT),
  fonttitle=\bfseries,
  fontupper=\small, 
  boxrule=0.5pt,
  arc=2mm,
  left=2mm, right=2mm, top=1mm, bottom=1mm,
  width=\textwidth 
]
You are a scientific peer-review hallucination detector.
\\[10pt]
\textbf{\#\# Task} \\
Determine whether the review sentence is hallucinated based strictly on the paper evidence.
\\[10pt]
\textbf{\#\# Definitions \& Constraints} \\
- A review claim is hallucinated if it is factually unsupported by, or incorrect relative to, the paper content. \\
- EXCLUSIONS: Do not evaluate tone (\textit{e.g.}, harshness). Do NOT treat subjective judgments, critiques, or opinions (\textit{e.g.}, whether novelty is high) as hallucinations. Focus strictly on factual grounding.
\\[10pt]
Reasoning steps: \\
1) Briefly verify key facts from \#Paper Evidence\# against the \#Review Sentence\#. \\
2) Check consistency between the paper's intent/context and the \#Review Sentence\#. \\
3) Conclude with the final JSON judgement.
\\[10pt]
---
\\[10pt]
\textbf{\#\# Inputs} \\
\textbf{\#\#\# PAPER EVIDENCE} \\
"""\{evidence\}"""
\\[5pt]
\textbf{\#\#\# REVIEW SENTENCE} \\
"""\{sentence\}"""
\\[10pt]
---
\\[10pt]
\textbf{\#\# Output Format} \\
Return a valid JSON object ONLY, using EXACTLY the following template:
\\[5pt]
\{
  "reasoning": "<Step-by-step analysis following the three reasoning steps above.>",
  "hallucination": <true or false>
\}
\\[10pt]
\textbf{\#\# Instructions} \\
- "true" = the sentence contains a factual claim that is hallucinated (unsupported or incorrect relative to the evidence). \\
- "false" = the sentence is factually supported by and consistent with the evidence, OR the sentence is purely a subjective judgment/critique. \\
- Output ONLY the JSON object. Do NOT include markdown tags like \texttt{```json}.
\end{tcolorbox}
\caption{Prompt template used for RA-LLM (CoT) in Task 1 Evaluation.}
\label{fig:prompt_RA-LLM_CoT}
\end{figure*}

\begin{figure*}
\centering
\begin{tcolorbox}[
  colback=gray!5,
  colframe=gray!180,
  title=Prompt Template for RA-LLM (Contrast),
  fonttitle=\bfseries,
  fontupper=\small, 
  boxrule=0.5pt,
  arc=2mm,
  left=2mm, right=2mm, top=1mm, bottom=1mm,
  width=\textwidth 
]
You are a scientific peer-review hallucination detector.
\\[10pt]
\textbf{\#\# Task} \\
Determine whether the review sentence is hallucinated based strictly on the paper evidence.
\\[10pt]
\textbf{\#\# Definitions \& Constraints} \\
- A review claim is hallucinated if it is factually unsupported by, or incorrect relative to, the paper content. \\
- EXCLUSIONS: Do not evaluate tone (\textit{e.g.}, harshness). Do NOT treat subjective judgments, critiques, or opinions (\textit{e.g.}, whether novelty is high) as hallucinations. Focus strictly on factual grounding.
\\[10pt]
Contrastive reasoning: \\
1) List the strongest evidence from \#Paper Evidence\# that supports the \#Review Sentence\#. \\
2) List the strongest evidence from \#Paper Evidence\# that contradicts the \#Review Sentence\# or shows missing facts. \\
3) Decide which side is stronger and output the final JSON judgement.
\\[10pt]
---
\\[10pt]
\textbf{\#\# Inputs} \\
\textbf{\#\#\# PAPER EVIDENCE} \\
"""\{evidence\}"""
\\[5pt]
\textbf{\#\#\# REVIEW SENTENCE} \\
"""\{sentence\}"""
\\[10pt]
---
\\[10pt]
\textbf{\#\# Output Format} \\
Return a valid JSON object ONLY, using EXACTLY the following template:
\\[5pt]
\{
  "supporting\_evidence": "<Strongest evidence from the paper that supports the review sentence.>",
  "contradicting\_evidence": "<Strongest evidence from the paper that contradicts the review sentence or shows missing facts.>",
  "hallucination": <true or false>
\}
\\[10pt]
\textbf{\#\# Instructions} \\
- "true" = the sentence contains a factual claim that is hallucinated (unsupported or incorrect relative to the evidence). \\
- "false" = the sentence is factually supported by and consistent with the evidence, OR the sentence is purely a subjective judgment/critique. \\
- Output ONLY the JSON object. Do NOT include markdown tags like \texttt{```json}.
\end{tcolorbox}
\caption{Prompt template used for RA-LLM (Contrast) in Task 1 Evaluation.}
\label{fig:prompt_RA-LLM_Contrast}
\end{figure*}

\begin{figure*}
\centering
\begin{tcolorbox}[
  colback=gray!5,
  colframe=gray!180,
  title=Prompt Template for Hallucination Detection,
  fonttitle=\bfseries,
  fontupper=\small, 
  boxrule=0.5pt,
  arc=2mm,
  left=2mm, right=2mm, top=1mm, bottom=1mm,
  width=\textwidth 
]
You are a scientific peer-review hallucination detector.
\\\\
\textbf{\#\# Task}\\
Determine whether the review sentence is hallucinated based strictly on the paper evidence.
\\\\
\textbf{\#\# Definitions \& Constraints}\\
- A review claim is hallucinated if it is factually unsupported by, or incorrect relative to, the paper content.\\
- EXCLUSIONS: Do not evaluate tone (\textit{e.g.}, harshness). Do NOT treat subjective judgments, critiques, or opinions (\textit{e.g.}, whether novelty is high) as hallucinations. Focus strictly on factual grounding.
\\\\
\textbf{\#\# Inputs}\\
\textbf{\#\#\# PAPER EVIDENCE}\\
"""\{evidence\}"""
 
\textbf{\#\#\# REVIEW SENTENCE}\\
"""\{sentence\}"""
\\\\
\textbf{\#\# Output Format}\\
Return a valid JSON object ONLY, using EXACTLY the following template:
 
\{
  "hallucination": <true or false>
\}
\\\\
\textbf{\#\# Instructions}\\
- "true" = the sentence contains a factual claim that is hallucinated (unsupported or incorrect relative to the evidence).\\
- "false" = the sentence is factually supported by and consistent with the evidence, OR the sentence is purely a subjective judgment/critique.\\
- Output ONLY the JSON object. Do NOT include markdown tags like \texttt{```json}.
\end{tcolorbox}
\caption{Prompt template used for hallucination detection with general-purpose LLM in Task 1 evaluation.}
\label{fig:prompt_hallucination_detection}
\end{figure*}

\begin{figure*}
\centering
\begin{tcolorbox}[
  colback=gray!5,
  colframe=gray!180,
  title=Prompt Template for Hallucination Type Classification,
  fonttitle=\bfseries,
  fontupper=\small, 
  boxrule=0.5pt,
  arc=2mm,
  left=2mm, right=2mm, top=1mm, bottom=1mm,
  width=\textwidth 
]
You are a scientific peer-review hallucination detector.
\\\\
\textbf{\#\# Task}\\
Given a review sentence that has ALREADY been judged as hallucinated, classify the hallucination TYPE using the provided taxonomy labels.
\\\\
\textbf{\#\# Definitions \& Constraints}\\
- The categories are mutually exclusive (MECE).\\
- You must rely strictly on the provided operational definitions in the HALLUCINATION LABELS, not just the label names.
\\\\
\textbf{\#\# Inputs}\\
\textbf{\#\#\# HALLUCINATION LABELS}\\
"""\{label\_space\}"""
 
\textbf{\#\#\# PAPER EVIDENCE}\\
"""\{evidence\}"""
 
\textbf{\#\#\# REVIEW SENTENCE}\\
"""\{sentence\}"""
\\\\
\textbf{\#\# Output Format}\\
Return a valid JSON object ONLY, using EXACTLY the following template:
 
\{
  "label": "<concept>"
\}
\\\\
\textbf{\#\# Instructions}\\
- The sentence is confirmed to be hallucinated. You MUST choose exactly ONE label corresponding to the "concept" field in the HALLUCINATION LABELS.\\
- Do NOT return "None" or invent new labels.\\
- If multiple labels seem plausible, choose the most specific one that aligns with the observable textual errors in the review sentence.\\
- Output ONLY the JSON object. Do NOT include markdown tags like \texttt{```json}.
\end{tcolorbox}
\caption{Prompt template used for hallucination detection with general-purpose LLM in Task 2 evaluation.}
\label{fig:prompt_hallucination_classification}
\end{figure*}

\begin{figure*}
\centering
\begin{tcolorbox}[
  colback=gray!5,
  colframe=gray!180,
  title=Prompt Template for Hallucination Span Localization,
  fonttitle=\bfseries,
  fontupper=\small, 
  boxrule=0.5pt,
  arc=2mm,
  left=2mm, right=2mm, top=1mm, bottom=1mm,
  width=\textwidth 
]
You are a scientific peer-review hallucination detector.
\\\\
\textbf{\#\# Task}\\
Given a list of review sentences, return exactly one sentence that exhibits the specific hallucination described in the target label.
\\\\
\textbf{\#\# Definitions \& Constraints}\\
- A claim is hallucinated if it is factually unsupported by, or incorrect relative to, the paper content.\\
- EXCLUSIONS: Do NOT treat subjective judgments, critiques, or opinions as hallucinations.\\
- You must strictly match the error in the sentence to the definition of the target label.
\\\\
\textbf{\#\# Inputs}\\
\textbf{\#\#\# TARGET HALLUCINATION LABEL}\\
"""\{label\_with\_definition\}"""
 
\textbf{\#\#\# PAPER EVIDENCE}\\
"""\{evidence\}"""
 
\textbf{\#\#\# REVIEW SENTENCES}\\
"""\{sentences\}"""
\\\\
\textbf{\#\# Output Format}\\
Return a valid JSON object ONLY, using EXACTLY the following template:
 
\{
  "result": []
\}
\\\\
\textbf{\#\# Instructions}\\
- Base your judgment ONLY on the PAPER EVIDENCE and the definition of the TARGET HALLUCINATION LABEL.\\
- If NO sentence matches the target label, output an empty array for the result field: []\\
- If a sentence matches, output an array containing EXACTLY ONE string: ["<The exact hallucinated sentence>"]\\
- If multiple sentences are hallucinated for this label, pick the single most definitive one. If uncertain, pick your best guess.\\
- Output ONLY the JSON object. Do NOT include markdown tags like \texttt{```json}.
\end{tcolorbox}
\caption{Prompt template used for hallucination span localization with generous-purpose LLM in Task 3 evaluation.}
\label{fig:prompt_hallucination_selection}
\end{figure*}

\end{document}